\documentclass{article}

\PassOptionsToPackage{numbers}{natbib}
\usepackage[final]{neurips_data_2023}

\usepackage[utf8]{inputenc} % allow utf-8 input
\usepackage[T1]{fontenc}  % use 8-bit T1 fonts
\usepackage{hyperref}    % hyperlinks
\usepackage{url}      % simple URL typesetting
\usepackage{booktabs}    % professional-quality tables
\usepackage{amsfonts}    % blackboard math symbols
\usepackage{nicefrac}    % compact symbols for 1/2, etc.
\usepackage{microtype}   % microtypography
\usepackage{xcolor}     % colors
\usepackage{times,latexsym}
\usepackage[T1]{fontenc}
\usepackage[utf8]{inputenc}
\usepackage{breakurl}
\usepackage{graphicx}
\usepackage{booktabs}
\usepackage{multirow}
\usepackage{tabularx}
\usepackage{float}
\usepackage{pifont}
\usepackage{makecell} 
\usepackage{amsmath}
\usepackage{array}
\usepackage{parskip}
\usepackage{rotating}
\usepackage{subfig}
\usepackage{chngpage}

\newcolumntype{L}[1]{>{\begin{varwidth}[t]{#1}}l<{\end{varwidth}}}

\title{RealKIE: Five Novel Datasets for Enterprise Key Information Extraction}

\author{%
 Benjamin Townsend\\
 Indico Data Solutions \\
 \texttt{ben@indico.io} \\
 \And
 Madison May \\
 Indico Data Solutions \\
 \texttt{madison@indico.io} \\
 \And
 Katherine Mackowiak \\
 Work Completed at \\
 Indico Data Solutions \\
 \texttt{katherinemackowiak@gmail.com} \\
 \AND
 Christopher M. Wells, Ph. D. \\
 Work Completed at \\
 Indico Data Solutions \\
 \texttt{physicistintheory@gmail.com} \\
}

\begin{document}

\newcommand{\cmark}{\ding{51}}
\newcommand{\xmark}{\ding{55}}
\newcommand*\rot{\rotatebox{90}}

\maketitle
\begin{abstract}
 We introduce RealKIE, a benchmark of five challenging datasets aimed at advancing key information extraction methods, with an emphasis on enterprise applications. The datasets include a diverse range of documents including SEC S1 Filings, US Non-disclosure Agreements, UK Charity Reports, FCC Invoices, and Resource Contracts. Each presents unique challenges: poor text serialization, sparse annotations in long documents, and complex tabular layouts. These datasets provide a realistic testing ground for key information extraction tasks like investment analysis and contract analysis. In addition to presenting these datasets, we offer an in-depth description of the annotation process, document processing techniques, and baseline modeling approaches. This contribution facilitates the development of NLP models capable of handling practical challenges and supports further research into information extraction technologies applicable to industry-specific problems. The annotated data, OCR outputs, and code to reproduce baselines are available to download at \href{https://indicodatasolutions.github.io/RealKIE/}{https://indicodatasolutions.github.io/RealKIE/}.
\end{abstract}

\section{Introduction}
\label{sec:intro}
The NLP community has a long history of producing and publishing benchmark datasets for information extraction tasks \cite{bib_conll03,bib_kleister,bib_ontonotes, bib_wnut, bib_cord, bib_cuad, bib_holt_chisholm, leivaditi2020benchmark, funaki2020contract, koreeda-manning-2021-contractnli-dataset}. Benchmarks like these have driven important advancements in Key Information Extraction (KIE) but save for the notable exceptions of \citet{bib_cuad} and \citet{bib_kleister}, they lack realism and do not adequately capture the complexity of tasks performed by knowledge workers in enterprise settings. The difficulties we intend to shed light on are:
\begin{itemize}
  \item poor document quality, leading to OCR artifacts and poor text serialization \cite{10.1145/1390749.1390753, artidigh20}
  \item sparse annotations within long documents that cause class imbalance issues \cite{10020873, park2022efficient, Li2021RethinkingNS}
  \item complex tabular layout that must be considered to discriminate between similar labels \cite{Koleva2022NamedER, Wang_2023, vanlandeghem2023document, 10.1145/2872518.2889386}
  \item data type variety (from simple dates to long-form clauses) that necessitates flexible models \cite{Wang2021CrossDomainCE}
\end{itemize}

These difficulties often arise due to processes upstream of the information extraction system. Since it is often impractical to tailor OCR, layout analysis, and text serialization to the business problem, models and machine learning systems that compensate for these artifacts would pose a significant boon to industry practitioners.

To this end, we present RealKIE, a benchmark of five document-level key information extraction datasets. Three document sources are novel as KIE tasks, while two expand upon the work of \citet{bib_kleister}. Included with the PDF documents are the full output of optical character recognition (OCR) and labelled spans indexed to that OCR output. Additionally, we share static train/test/validation splits to benefit extensibility and reproducibility. In all cases, the fields extracted are representative of data extraction tasks for a specific industry vertical. We hope that these new benchmarks will spark research into novel approaches to information extraction in real-world settings and drive the development of models and methods directly applicable to industry problems. 

We release the labels and OCR under the CC-BY-NC 4.0 license.

\section{Dataset Descriptions}
\label{sec:datasets}

% This section is a summary of the datasets that compose RealKIE. Each subsection contains a description of the documents, example elements from the full sequence labeling schema, and summary statistics. 

Each task follows the same high-level format, taking a visually-rich document \cite{Wang_2023} as input and producing labeled spans from the document. While RealKIE makes no strong assertions about the input representation to the model, this may include a combination of: a linear sequence of tokens, token position information, and a pixel representation of each page.

We define labeled spans as character start and end indices (relative to the provided OCR text of the document) along with a corresponding field name.  While our baselines in Section \ref{sec:results} use a token classification formulation, we impose no constraints on how these outputs are derived.

The format of this task is broadly similar to Named Entity Recognition (NER) but with two main differences \cite{bib_conll03}. 
\begin{itemize}
  \item \textbf{Field Specificity} - NER fields are usually broad categories such as "Person Name". For RealKIE, we are also interested in the role of the entities in the document. For example, a person name could be labeled as "Trustee Name", "Donor Name" or not labeled at all.
  \item \textbf{Label Length} - NER labels are usually short spans. While many of our labels are short, we also have 15 paragraph-level fields across our datasets.
\end{itemize}
Section \ref{sec:dataset_compare} shows a full comparison to existing datasets.

For all tasks, the complete list of fields and their label counts can be seen in Appendix \ref{sec:tables-appendix}.

All documents used in RealKIE are required to be made public by regulatory requirements intended to provide transparency and were public prior to the construction of RealKIE.  They are made available in part to invite the scrutiny of the public and ensure practices are held to an ethical standard.  In addition, these documents focus on the organizations they concern and do not include personal details that would typically be considered sensitive (SSNs, bank information, medical information, demographic data, or personal contact information).

\subsection{SEC S1 Filings}
This dataset consists of 322 labeled S1 filings sourced from the Security and Exchange Commission's (SEC) EDGAR data store \cite{SECgovPr80:online}. The SEC requires domestic issuers to file an S1 prior to publicly offering new securities, most commonly as part of an initial public offering. While these documents are all required to contain certain sections (e.g., risk factors and the details of the securities offered) and are known as registration \emph{forms}, there is a high degree of variability in the document content and presentation. While some filings are born-digital PDFs, others have been scanned before uploading. Furthermore, these docs are often lengthy, and the labels are sparsely scattered throughout the document, leaving many full pages unlabeled. These characteristics make S1 Filings an ideal representation of a typical enterprise knowledge worker task.

The labeling schema represents the activities of an investment analyst assessing whether to invest in a given offering, for example, by extracting risk factor statements. We have also included header fields for key sections like the prospectus summary and the detailed description of the securities.

\subsection{US Non-Disclosure Agreements (NDA)} \label{sec:kleister_nda}
This dataset contains 439 non-disclosure agreements submitted to EDGAR as a part of various required filings \cite{SECgovPr80:online}. The raw documents were thoroughly presented in \citet{bib_kleister}, so we focus on our contributions. We include a similar label schema to the original Kleister-NDA dataset, omitting the term field \cite{bib_kleister}. This schema covers fields extracted in a legal setting: the parties involved, the effective date, and the jurisdiction of the contract. Unlike the original annotations, we provide manually labeled text span annotations referenced against the document text. While the original documents were in an HTML format, we use rendered PDFs shared as part of the Kleister NDA dataset \cite{bib_kleister}. Though we annotate only a trio of fields, this task proves challenging due to label sparsity. 

\subsection{UK Charity Reports} \label{sec:kleister_charity}
 
This dataset contains 538 public annual reports filed by charities in the UK with the UK Charity Commission. Our document set contains partial overlap with Kleister-Charities documents \cite{bib_kleister}.  Similar to those in Section \ref{sec:res_con}, these documents are lengthy, and while they all carry similar information, formatting varies significantly between documents. As such, they represent the types of documents a knowledge worker might scour for details in an audit or diligence setting. 
 As in the NDA dataset in Section \ref{sec:kleister_nda}, this dataset was first compiled and modeled in \citet{bib_kleister}, so we focus on our contributions. The schema we have applied to these documents extends that of Kleister-Charities \cite{bib_kleister}. We include fields that capture information about the charity's activities, including named charity events and the names and roles of trustees. These fields are particularly challenging due to their mixed data types and presentations within the document.

\subsection{FCC Invoices}
This dataset consists of 370 labeled invoices that contain cost information from television advertisements placed by political campaigns. These Federal Communication Commission (FCC) filings are required to be made public as part of U.S. political campaign disclosure policies \cite{AboutFCC18:online}.

As with most invoices, they have a mixture of:
\begin{itemize}
  \item document-level metadata, e.g., the agency placing the ad and the client on whose behalf it is being placed
  \item line-level information, e.g., the start/end dates of a billing period and the rate per spot
  \item summary information, e.g., gross and net amounts invoiced
\end{itemize}

In some sense, these documents are the most structured of all the documents presented here, as documents from the same broadcasters often share the same layout. However, the presentation varies considerably between broadcasters. In addition, table nesting and the format of certain data elements mean that understanding the spatial layout of the document is critical to extract key information correctly. This dataset represents the activities of a knowledge worker in accounts payable or accounts receivable tasked with extracting key details from invoices.

\subsection{Resource Contracts}\label{sec:res_con}
This dataset consists of 198 labeled legal contracts specifying the details of agreements to explore for and exploit natural resources. These contracts specify the details of the geography, the dates of project phases, revenue-sharing agreements, and tax laws. The documents have been sourced from the Resource Contracts Online Repository, an open repository of global mining and petroleum contracts \cite{bib_resource_contracts}. 

These documents are challenging for a variety of reasons. While they all contain roughly the same information, their formats are highly varied. They span many decades and the spectrum of visual quality, including text within images, machine text, and handwriting. As such, there is a high degree of OCR ambiguity, as discussed in Section \ref{sec:dataset-analysis}. 

Our labeling schema differs from the originals provided by \citet{bib_resource_contracts} and is meant to represent the activities of an attorney performing contract diligence. At a high level, the annotated data elements fall into three categories:
\begin{itemize}
  \item preamble fields, e.g., the named parties to a given contract or the date it was signed
  \item header fields, i.e., the headings of key sections, meant to simplify navigation in and through a highly self-referential document
  \item clause fields, e.g., the obligations of a contractor with respect to environmental protections
\end{itemize}

\section{Document Processing} \label{sec:processing}

Each document enters our document processing pipeline as a PDF. Each page of the PDF is rendered as an image and processed by an OCR engine. Some documents are born-digital or partially digital PDFs, but for consistency, every document goes through an OCR process. The OCR files, images, and original files are all shared as part of the dataset. Any documents with exactly duplicate text were removed from the dataset.

We use two different pipelines to process the documents. For the OmniPage pipeline, we use OmniPage to both OCR and convert the PDF files to PNG \cite{kofax_omnipage_2023}. For the Azure Read OCR Pipeline, we use Azure Computer Vision Read API (version 2021-04-12) \cite{azure_read_api_2023} to OCR the PDF and then PyPDFium to convert the files to PNGs \cite{korobov2023pypdfium2}. In both cases, rotation and de-skewing are applied according to the outputs of the OCR engines. 

OmniPage was used for all datasets except Resource Contracts. The Resource Contracts files include shading and partial occlusion from poor-quality scans and handwriting, which are handled better by Azure's Read OCR.
% An Example can be seen in Figure \ref{fig:resource-constracts-scan-1}

This simple document processing workflow plays an important role in our dataset preparation process. By making this OCR data publicly accessible, we hope to establish a level playing field on which to compare subsequent works. Without standard OCR, it can be challenging to decouple improvements due to data pre-processing and improvements due to modeling advances.

We randomly split the documents into train, test, and validation splits, allocating 20\% each for test and validation and the remainder for train. 

The size of the documents in the S1 dataset results in practical challenges for benchmarking. To make our datasets as accessible as possible, we have segmented the documents at the page level. Pages without labels have been removed at random to reduce the dataset's overall size. This segmentation was done after constructing train/test/val splits at the document level to avoid data contamination. The S1 dataset with labels and OCR is available in both page and document level formats. 

% \begin{figure}
%   \centering
%   \includegraphics[width=0.5\textwidth]{resource-contracts-scan-1.png}
%   \caption{An example of a poor-quality scan from resource-contracts. OmniPage fails to read any text along the vertical stripe and handles it as a 2 column document. Azure Read recognizes the text almost perfectly.}
%   \label{fig:resource-constracts-scan-1}
% \end{figure}

\section{Description of Annotation Task} \label{sec:annotation}

 Professional annotators used a commercial annotation interface to label these documents \cite{IndicoData_2023}. The process consisted of three main phases: initial annotation, model-assisted annotation, and quality review. More details of the annotation process can be found in Appendix \ref{sec:annotation-appendix}.

% The number of errors detected by the model for each dataset can be seen in Table \ref{tab:model_qa_table}

% \begin{table}[]
%   \centering
%   \begin{tabular}{l|c|c}
%     Dataset Name & Missed & Extra \\
%     \hline
%     NDA & 226 & 25 \\
%     Charities & 3930 & 739\\
%     S1 & 3129 & 362\\
%     Resource Contracts & 7959 & 322 \\
%     FCC Invoices & 8166 & 5035 \\
%   \end{tabular}
%   \caption{The counts of the types of errors detected by the model-based dataset QA process defined in Section \ref{model_qa_section}}
%   \label{tab:model_qa_table}
% \end{table}

\section{Dataset Analysis}
\label{sec:dataset-analysis}

In this section, we analyze our datasets to demonstrate their effectiveness in representing the challenges described in Section \ref{sec:intro}.

\subsection{Complex Layouts and Poor OCR Quality}

\begin{table}
\centering
\begin{tabular}{l|l|l|lll|}
\cline{2-6}
                     & OCR Confidence      & \multicolumn{1}{c|}{Text Linearity} & \multicolumn{3}{c|}{Table Frequency}                              \\
                     & \multicolumn{1}{c|}{Mean} & \multicolumn{1}{c|}{Mean}      & \multicolumn{1}{c}{Docs} & \multicolumn{1}{c}{Pages} & \multicolumn{1}{c|}{Labeled Tables} \\ \hline
\multicolumn{1}{|l|}{NDA}        & 100\%           & 99.0                & 0\%           & 0\%            & 0\%                   \\ \hline
\multicolumn{1}{|l|}{Charities}     & 98.2\%          & 91.4                & 87.0\%          & 33.9\%          & 22.7\%                 \\ \hline
\multicolumn{1}{|l|}{FCC Invoices}    & 94.2\%          & 80.6                & 93.2\%          & 74.6\%          & 45.6\%                  \\ \hline
\multicolumn{1}{|l|}{Resource Contracts} & 94.7\%          & 95.8                & 79.8\%          & 3.3\%          & -                    \\ \hline
\multicolumn{1}{|l|}{S1}         & 100\%           & 99.1                & 3.1\%          & 3.1\%           & 66.1\%                 \\ \hline
\end{tabular}
\vspace{0.5em} % Not sure why these are so close otherwise
\caption{Measures indicating the OCR quality and Layout style of the documents in the datasets. OCR confidence is the mean character recognition confidence returned by OmniPage, which is correlated with the accuracy of character extraction. We have measured text linearity by calculating the per-page Levenshtein similarity ratio between the OCR text and the same text re-ordered with a top to bottom, left-to-right reading order. Finally, table frequency is reported, where the presence of a table is based on Omnipage's table detection. In addition, where possible, we report the percentage of tables that include at least one label. }
\label{tab:layout-and-quality}
\end{table}

In this section, we will analyze the layout complexity and OCR quality of our datasets. Table \ref{tab:layout-and-quality} shows three measures of layout complexity and OCR quality: OCR Confidence, Text Linearity, and Table Frequency. From this, we see that the NDA and S1 documents are the least complex of the five datasets, with high linearity scores, few tables, and perfect OCR confidence scores.

\begin{figure}
  \centering
  \includegraphics[width=1\linewidth]{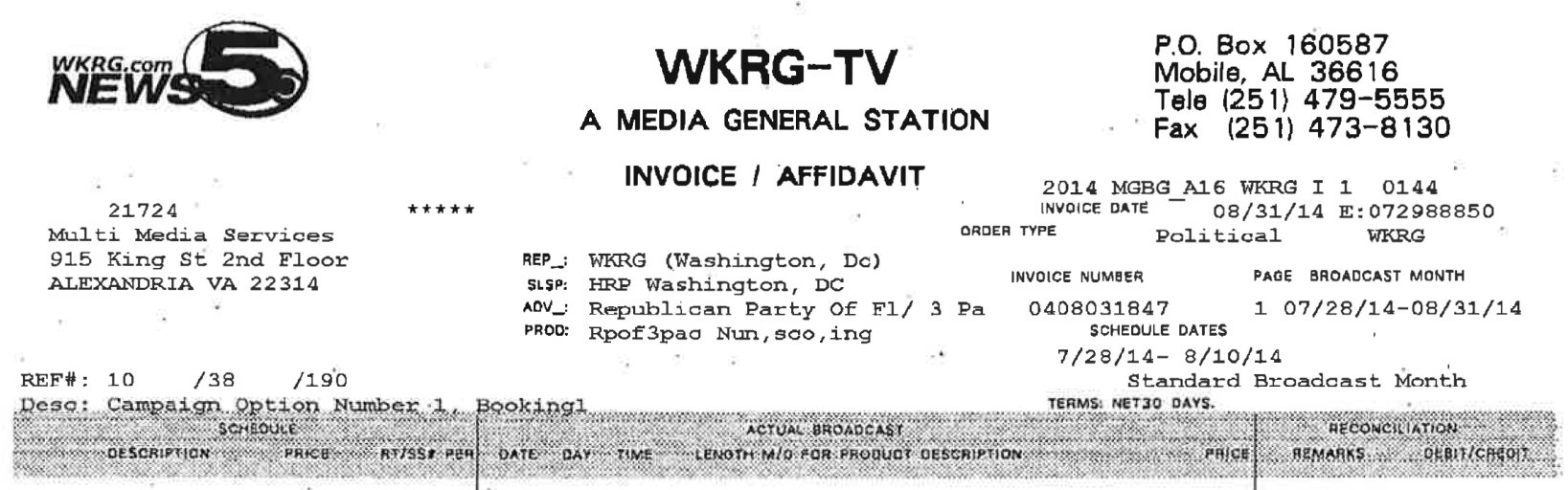}
  \caption{This snippet of an FCC invoice is an example of reading order ambiguity and character recognition ambiguity. There are many equally correct ways to serialize this content. This characteristic is referred to as inherent reading order ambiguity.
  The bottom lines illustrate lower OCR confidences, indicating character recognition ambiguity. We can see that processes applied to this document, likely being printed and then scanned, have introduced some corruption of letters with "PRICE" reading as "PRICB" and "SCHEDULE" as "SCNEOULE".
  }
  \label{fig:ambiguous-reading}
\end{figure}

Low text linearity scores, as seen in the Charities, FCC Invoices, and Resource Contracts datasets, may have two primary causes:

\begin{itemize}
  \item Inherent reading order ambiguity - Cases where there is no well-defined order to read a document. An example is shown in Figure \ref{fig:ambiguous-reading}.
  \item OCR reading order failures - Cases where a reading order is well-defined, but the OCR engine has incorrectly interpreted the document - removing critical information in the process. 
% One example of this is found in definition pages. Often, terms in definition pages are left-justified and definitions are right-justified, in some cases an OCR engine may incorrectly read this as a two column page. This text serialization failure makes inferring associations between keys and values more difficult when analyzing the document text alone. 
\end{itemize}

FCC Invoices exhibit low text linearity due to a large number of complex tables and dense areas of key-value information. Both Charities and Resource Contracts show indications of high-layout complexity. 

OCR confidence is a directional indicator of character recognition ambiguity. Figure \ref{fig:ambiguous-reading} shows an example of lower confidence OCR.
FCC Invoices and Resource Contracts have the lowest OCR confidence scores, indicating more frequent character-level OCR errors.

Table presence contributes to overall document complexity; however, the impact on the overall benchmark is higher if the tables contain labeled spans. A significant portion of the tables in Charities, FCC Invoices, and S1s contain labeled spans. In the case of Charities and FCC invoices, interpreting table structure is critical for effectively solving the task.

\begin{figure}
  \centering
  % \subfloat[Charities]{\includegraphics[width=0.48\linewidth]{charities_table.png}}\hfill
  % \subfloat[FCC Invoices]{
  \includegraphics[width=0.70\linewidth]{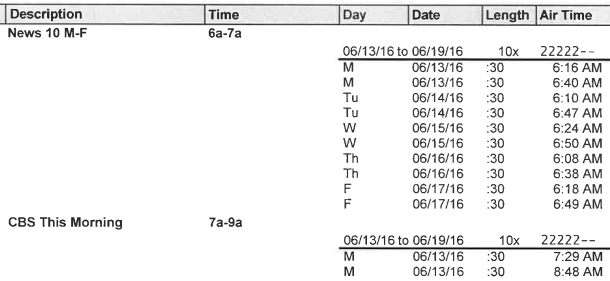}
  % }
  \caption{Part of a table from the FCC Invoices dataset. In Table \ref{tab:layout-and-quality}, this would simply show as a table. However, it contains features that significantly increase modeling difficulty compared to a typical table structure. For example, the slots per day indicator "22222--" is directly under the Air Time header but does not relate to it. Similarly, the date range values of the outer table are merged left across another labeled "Day" header. These complications vary significantly between different broadcasters.}
  \label{fig:table-compare}
\end{figure}

Not all elements of document complexity are captured by these metrics. For example, the tables in FCC invoices are far more complex than the ones in Charities. However, both are counted equally in Table \ref{tab:layout-and-quality}. Figure \ref{fig:table-compare} shows an example of how tables can vary in complexity.

Overall, these five datasets span a wide range of layout and OCR-related complexities. Further discussion and comparison to previous work are included in Section \ref{sec:dataset_compare}

\subsection{Sparse Annotations}

When training transformer models on long documents such as ours, it is necessary to chunk the document into smaller sections that fit within the context size of the model \cite{Dai2019TransformerXLAL}. Table \ref{tab:class_imbalance} shows two kinds of imbalance within the datasets: an imbalance between chunks of the document that contain labels and do not, and an imbalance between the labels within the document. 

For imbalances between labeled chunks and unlabeled chunks, we can see that this ranges between 81.8\% empty chunks for NDA and 0\% empty chunks for FCC Invoices. For NDA, this means that, with the RoBERTa tokenizer and a chunk size of 512 tokens, only 18.2\% of chunks contain any labels. This empty-chunk rate is an important form of sparsity to consider when modeling these datasets.

Including the background class, we can see that the token class imbalance is as high as 12000:1 for Charities. Excluding background tokens gives us a more reasonable 160:1 imbalance for the same dataset. This drop shows that sparsity is the main driver of class imbalance.

\begin{table*}[]
  \centering
  \small{
\begin{tabular}{lr|r|r|}
\cline{2-4}
          &\multicolumn{1}{|l|}{}            & \multicolumn{2}{c|}{Class Imbalance} \\
          \cline{3-4}
    & \multicolumn{1}{|r|}{\% Chunks Without Labels} & Including Background & Excluding Background \\
   
\hline
     \multicolumn{1}{|l|}{Charities} &         25.00 &       12364.52 &                159.40 \\
        \multicolumn{1}{|l|}{NDA} &         81.82 &       3007.84 &                 10.64 \\
     \multicolumn{1}{|l|}{S1} &         50.00 &       7679.23 &                882.13 \\
\multicolumn{1}{|l|}{Resource Contracts} &         78.16 &       17496.17 &                150.76 \\
   \multicolumn{1}{|l|}{FCC Invoices} &         0.00 &       1082.43 &                 67.68 \\
\hline
\end{tabular}
}

  \caption{Showing the percentage of chunks without labels and the maximum class imbalances with and without the background class. Maximum class imbalance is the ratio between the number of labeled tokens in the most frequent and least frequent classes. Chunks are computed using the RoBERTa tokenizer and a chunk size of 512 tokens. As a result of long documents with sparse labels, the imbalance between labels and the background class is often severe.}
  \label{tab:class_imbalance}
\end{table*}

\subsection{Data Type Variety}

Table \ref{tab:label_types} shows the label diversity in our datasets. There are 16 different data types extracted in our datasets. Common data types, such as Date, occur in every dataset, whereas other types, such as City and Country Name, only occur in a small subset. Data types such as "Clause" and "Phrase" are diverse: each of the 19 underlying fields represents a unique type of extraction with its own considerations.

The variety of data types presents a number of challenges and opportunities. For example, clause-based label spans may be longer than a model's context size, and header labels may benefit from additional information, such as formatting.

\begin{table}
  \centering
    \begin{tabular}{l|rrr}
    Data Type         & \multicolumn{1}{l}{Instance Count} & \multicolumn{1}{l}{Field Count} & \multicolumn{1}{l}{Dataset Count} \\
    \hline
    Address        & 2910                & 5               & 2               \\
    City         & 1428                & 2               & 1               \\
    Clause        & 32766               & 15               & 3               \\
    Company Name     & 12640               & 9               & 4               \\
    Country Name     & 432                & 1               & 1               \\
    Date         & 30063               & 6               & 5               \\
    Header        & 3239                & 12               & 2               \\
    Misc         & 13190               & 1               & 1               \\
    Monetary Value    & 24661               & 11               & 3               \\
    Other Name      & 660                & 3               & 2               \\
    Person / Company Name & 1495                & 2               & 2               \\
    Person Name      & 11208               & 7               & 3               \\
    Phrase        & 18509               & 4               & 3               \\
    Post Code       & 1339                & 2               & 1               \\
    Title         & 1813                & 1               & 1               \\
    Year         & 6354                & 1               & 1               
    \end{tabular}
  \caption{This table shows the different data types of labels in our datasets, the total count of instances of each type, the number of labels of that type, and the number of datasets in which this type appears.}
  \label{tab:label_types}
\end{table}

\section{Comparison to Existing Datasets}

\label{sec:dataset_compare}
There are numerous document KIE datasets in the literature; a comparison is shown in table \ref{tab:dataset_compare}. 

SROIE and CORD are datasets of receipts. SROIE captures a simple schema of just four labels. CORD expands on this for a much larger schema of 30 fields \cite{SROIE, bib_cord}. These are datasets of very short documents (receipts) with complex layouts. However, due to the context limits of models such as LayoutLM-v3 \cite{huang2022layoutlmv3}, it is useful to have datasets such as our FCC Invoices dataset, which captures very complex layouts and exceeds the context limits of such models. We anticipate that this dataset will be a valuable benchmark for work involving improved chunking or longer-context layout models.

Kleister NDA, Kleister Charities, and DeepForm are existing information extraction datasets that pair documents with various fields and data types \cite{DeepForm2020, bib_kleister} and represent useful tasks in industry. However, due to the way these datasets were constructed, they do not contain the locations of the values to be extracted (span labels). As described in \citet{townsend2021doc2dict}, several challenges are associated with modeling datasets without span labels. One important example is that ambiguity is introduced when converting the labels into labeled tokens, as many modeling approaches require. This can make it difficult to decouple modeling advances from improvements to this mapping process. Providing OCR data and label spans also simplifies the evaluation process, as without information about where each extraction was found in the document, it can be difficult to assess model performance. Extensive postprocessing may be necessary to determine, for example, whether the extraction "26 January 2024" is equivalent to the expected result "01/26/2024".

While our Charities dataset is based on the same document source as Kleister Charities \cite{bib_kleister}, the RealKIE schema is more extensive, containing 28 fields instead of 8 and covering a mixture of different data types including names, dates, monetary values, and paragraph descriptions. This does come at the expense of the size, with RealKIE Charities containing only 538 documents compared to 2778 in the Kleister dataset.

CUAD is a dataset of contracts with expert legal clause annotations \cite{bib_cuad} and is complementary to the datasets presented here. CUAD contains long documents with a complex label schema relevant to contract review problems in industry. Our S1 and Resource Contracts datasets include some clause fields, however none of our datasets are clause level-only. 

SIREX and FUNSD are datasets with simple KIE schemas and a focus on entity linking \cite{jain-etal-2020-scirex, FUNSD}. SIREX is a dataset of machine learning papers with Dataset, Method, Task, and Metric fields as well as relationships between them \cite{jain-etal-2020-scirex}. FUNSD is a form dataset containing question, answer, header, and "other" labels. Complexity in both datasets stems from the variety of values that a particular field may capture. Both datasets focus on entity linking, which is orthogonal to our work.

In addition to the KIE datasets discussed there are a number of datasets which apply different tasks to related sets of documents. For example, FinTOC is a table of contents extraction task on financial disclosures \cite{fintoc}. ContractNLI is an Natural Language Inference task applied to NDAs from EDGAR, the same original source as our NDA dataset \cite{koreeda2021contractnli}. Finally, LEDGAR is a clause classification dataset from a range of different SEC filings from EDGAR \cite{tuggener2020ledgar}. In all cases these datasets use text-based versions of the data, bypassing the need for an OCR pipeline and the noise that this introduces. Some aspects of the FinTOC and LEDGAR tasks are captured by our SEC and S1 datasets with the inclusion of header and clause type fields but overall these are distinct from the primarily KIE focus of our RealKIE datasets.

% Each of the datasets surveyed has unique characteristics that make it valuable for KIE research, the value of the our datasets is the variety and complexity which we believe is well representative of some of the challenges observed in industrial applications of KIE.

\begin{table*}
  \centering
  \begin{tabular}{l|c|c|c|c||c|c|c|c|c}
   Name            & \rot{OCR Artifacts} & \rot{Varied Data Types} & \rot{Long Documents} & \rot{Complex Layout} & \rot{Manually Labeled} & \rot{Span Labels} & \rot{Num Fields} & \rot{Num Documents} & \rot{Num Annotations} \\
    \hline
    Charities         & \cmark       & \cmark        & \cmark     & \cmark     & \cmark         & \cmark   & 28     & 538      & 33366   \\
    NDA            & \cmark       & \cmark        & \cmark     & \xmark     & \cmark         & \cmark   & 3      & 439      & 1799   \\
    S1             & \cmark       & \cmark        & \xmark     & \xmark     & \cmark         & \cmark   & 24     & 13097     & 41833   \\
    FCC Invoices        & \cmark       & \cmark        & \cmark     & \cmark     & \cmark         & \cmark   & 11     & 370      & 82640   \\
    Resource Contracts     & \cmark       & \cmark        & \cmark     & \cmark     & \cmark         & \cmark   & 23     & 198      & 9005   \\
    \hline
    Kleister NDA \cite{bib_kleister}       & \xmark       & \cmark        & \cmark     & \xmark     & \xmark         & \xmark   & 4      & 540      & 2160     \\
    Kleister Charities \cite{bib_kleister}    & \cmark       & \cmark        & \cmark     & \cmark     & \xmark         & \xmark   & 8      & 2778     & 21612    \\
    Deepform \cite{DeepForm2020}         & \xmark         & \cmark        & \cmark     & \xmark     & \xmark         & \xmark   & 4      & 9018     & $\approx$36072 \\
    SROIE \cite{SROIE}          & \cmark       & \cmark        & \xmark     & \cmark     & \cmark         & \cmark   & 4      & 973      & 3892     \\
    CORD v2 \cite{bib_cord}          & \cmark       & \cmark        & \xmark     & \cmark     & \cmark         & \cmark   & 30     & 1000     & 13515  \\
    FUNSD \cite{FUNSD}           & \cmark       & \xmark        & \xmark     & \cmark     & \cmark         & \cmark   & 4      & 199      & 9743 \\
    SIREX \cite{jain-etal-2020-scirex}           & \xmark       & \xmark        & \cmark     & \xmark     & \cmark         & \cmark   & 5      & 438      & 156931 \\
    CUAD \cite{bib_cuad}           & \xmark       & \xmark        & \cmark     & \xmark     & \cmark         & \xmark   & 41     & 510      & 13101 \\

  \end{tabular}
  \caption{A comparison of the currently available datasets and how they compare to the goals of RealKIE. Complex layout is defined as something other than simple linear text, where elements of the non-linear text (such as a table) are relevant to solving the task.
  }
  \label{tab:dataset_compare}
\end{table*}

\section{Baseline Procedure and Results} 
\label{sec:experiments}

For our baselines, we finetune four pre-trained transformers with a token-classification formulation. Code to reproduce our baselines is available at \href{https://indicodatasolutions.github.io/RealKIE/}{https://indicodatasolutions.github.io/RealKIE/} For each model and dataset combination, we ran a Hyperband Bayesian hyper-parameter search until 100 models had trained \cite{li2018hyperband, wandb}. We then select the model with the highest validation set F1 and report test set F1. We evaluate RoBERTa-base, DeBERTa-v3-base, LayoutLM-v3-base and Longformer-base \cite{liu2019roberta, he2021debertav3, huang2022layoutlmv3, Beltagy2020Longformer}. Details for these models can be found in Table \ref{tab:model_info}.

When training on long documents with sparse labels, it is necessary to chunk the document into lengths determined by the context size of the model \cite{Dai2019TransformerXLAL}. It can be helpful to undersample chunks without labels to improve recall and stabilize the loss \cite{Li2021RethinkingNS}. For our baselines, we include an option to undersample negative chunks to a target ratio of labeled chunks to chunks without labels. This ratio is included in our hyperparameter search.

\subsection{Hardware and Environmental Impact}
 Running the baselines took approximately 16 GPU-Months. Following the method from \citet{lacoste2019quantifying} resulted in an estimated equivalent CO2 of 286 kg. The baselines were run on a combination of local 2080TI GPUs and T4 instances from AWS. The authors believe that the impact is justified by producing reliable baselines to be re-used in future work without the necessity for full reproduction. Code for reproducing baselines is available at \href{https://indicodatasolutions.github.io/RealKIE/}{https://indicodatasolutions.github.io/RealKIE/}.

\begin{table*}[h]
\centering
\small{
\begin{tabular}{l|c|c|c}
Model Name &2D Position & Max Length & \# Parameters\\
\hline
RoBERTa Base   \cite{liu2019roberta}    & \xmark & 512 & 125M \\
DeBERTa-v3 Base \cite{he2021debertav3}    & \xmark & 512 & 184M \\
LayoutLM-v3 Base \cite{huang2022layoutlmv3}  & \cmark & 512 & 133M \\
Longformer Base \cite{Beltagy2020Longformer} & \xmark & 4096 & 149M \\
\end{tabular}
}
\caption{Baseline Model Info}
\label{tab:model_info}
\end{table*}

\subsection{Results} \label{sec:results}

\begin{table*}[]
\centering\small{
\begin{tabular}{|c|c|c|c|c|}
\hline
\textbf{Dataset} & \textbf{Base Model} & \textbf{Test Macro F1} & \textbf{Val Macro F1} \\
\hline
\hline\multirow{4}{*}{Charities} & Longformer Base & 58.1 & 59.9 \\
\cline{2-4}
 & \textbf{LayoutLM-v3 Base} & \textbf{63.6} & 62.6 \\
\cline{2-4}
 & DeBERTa-v3 Base & 61.3 & \textbf{64.2} \\
\cline{2-4}
 & RoBERTa Base & 57.6 & 61.6 \\
\cline{2-4}
\hline\multirow{4}{*}{FCC Invoices} & Longformer Base & 67.3 & 74.8 \\
\cline{2-4}
 & LayoutLM-v3 Base & 68.3 & 75.6 \\
\cline{2-4}
 & \textbf{DeBERTa-v3 Base} & \textbf{69.2} & \textbf{76.4} \\
\cline{2-4}
 & RoBERTa Base & 66.5 & 73.1 \\
\cline{2-4}
\hline\multirow{4}{*}{NDA} & Longformer Base & 81.0 & \textbf{84.2} \\
\cline{2-4}
 & LayoutLM-v3 Base & 80.7 & 82.0 \\
\cline{2-4}
 & \textbf{DeBERTa-v3 Base} & \textbf{83.7} & 82.8 \\
\cline{2-4}
 & RoBERTa Base & 81.5 & 82.8 \\
\cline{2-4}
\hline\multirow{4}{*}{Resource Contracts} & Longformer Base & 45.5 & 44.9 \\
\cline{2-4}
 & LayoutLM-v3 Base & 41.8 & 45.0 \\
\cline{2-4}
 & \textbf{DeBERTa-v3 Base} & \textbf{45.6} & \textbf{46.1} \\
\cline{2-4}
 & RoBERTa Base & 40.9 & 44.0 \\
\cline{2-4}
\hline\multirow{4}{*}{S1} & Longformer Base & 82.6 & \textbf{83.5} \\
\cline{2-4}
 & \textbf{LayoutLM-v3 Base} & \textbf{83.5} & 83.4 \\
\cline{2-4}
 & DeBERTa-v3 Base & 81.8 & 81.1 \\
\cline{2-4}
 & RoBERTa Base & 81.7 & 82.6 \\
\cline{2-4}
\hline
\end{tabular}

}
\caption{Test set Macro F1 by dataset and by model. Models are selected based on the best validation F1 for each dataset and base model.  Metrics were computed per instance rather than per document. For example, if there were 3 instances of the field "Net Amount Due" within a document, a false positive was recorded for missing any of these instances, even if the information was redundant. A true positive is only recorded if a model extracts the correct content from the correct location in the document.}
\label{tab:main_results}
\end{table*}

Our baseline results (table \ref{tab:main_results}) indicate that DeBERTa-v3 is the highest-performing overall model and provides a strong baseline across the five datasets.

LayoutLM-v3 is the best performer on Charities and S1. The most consistent benefits of LayoutLM-v3 are found on the paragraph-level fields, where improvements as high as 19 F1 are observed. Counter-intuitively, while LayoutLM-v3 demonstrates improvement over text-only models on paragraphs, it does not consistently outperform the text-only models on tables. Margins are consistently small between DeBERTa-v3 and LayoutLM-v3 for tabular fields.

As Longformer is a RoBERTa derivative with a larger context size \cite{Beltagy2020Longformer}, comparing these two models hints at whether long-term context is useful for solving these information extraction tasks.  As expected, Longformer outperforms RoBERTa on 4/5 tasks, indicating that several fields are likely to benefit from long-context

Please refer to Appendix \ref{sec:tables-appendix} for a detailed breakdown of field-level metrics.

\subsection{Limitations}

\textbf{Single Annotations} - These datasets were labeled by a single annotator with a partially model-assisted review. Consequently, we do not provide measures such as inter-annotator agreement, which may be helpful for contextualizing baseline results.

\textbf{Baseline Diversity} - We chose to focus our baselines on the token-classification formulation; this notably omits the class of OCR-free methods \cite{kim2022ocrfree}. While the datasets are applicable to this form of model, we omit them from our baselines to allow us to standardize on a span-based metric formulation.

\textbf{English Only} - This benchmark focuses on datasets containing only English documents. Non-English or mixed-language business documents represent another important set of challenges not covered by this work.

\textbf{Dataset Size} - RealKIE datasets vary in size from 198 documents to 538 documents, which is smaller than \citet{bib_kleister} and other similar benchmarks.

\section{Conclusions}
 \label{sec:conclusions}
In this paper, we have introduced RealKIE, a benchmark of five document datasets. These documents and the associated tasks are faithful representations of many of the challenges encountered when automating data extraction:
\begin{itemize}
  \item poor document quality, leading to OCR artifacts and poor text serialization
  \item sparse annotations within long documents that cause class imbalance issues
  \item complex tabular layout that must be considered to discriminate between similar labels
  \item varied data types to be extracted: from simple dates and prices to long-form clauses
\end{itemize}

Models or frameworks that can improve upon the benchmarks presented here (by being robust to these common difficulties) would represent a major step forward in real-world information extraction technologies. We hope that RealKIE will be a reusable test bed for such advances.

\section{Acknowledgments}
We would like to acknowledge, by name, the substantial effort expended by our labeling team to produce high-quality labels for these complex datasets; many thanks to Ash Sloban, Jay Morgan, Lavi Sanchez, Melissa Cano, Sarah Magnant, Sidney More, Mackenzie Dwyer, and Donna Waltz.

\clearpage

\bibliography{neurips.bib}
\bibliographystyle{unsrtnat}

% \onecolumn
\clearpage

\appendix

\newpage
\section*{NeurIPS Paper Checklist}

\begin{enumerate}

\item {\bf Claims}
  \item[] Question: Do the main claims made in the abstract and introduction accurately reflect the paper's contributions and scope?
  \item[] Answer: \answerYes{} % Replace by \answerYes{}, \answerNo{}, or \answerNA{}.
  \item[] Justification: The goal of the paper is to present five datasets that capture the realistic challenges outlined in our introduction. Through our analysis, we have demonstrated how the datasets provided cover these challenges. 

\item {\bf Limitations}
  \item[] Question: Does the paper discuss the limitations of the work performed by the authors?
  \item[] Answer: \answerYes{}{} % Replace by \answerYes{}, \answerNo{}, or \answerNA{}.
  \item[] Justification: We have outlined several key limitations of our datasets in the Limitations section.

\item {\bf Theory Assumptions and Proofs}
  \item[] Question: For each theoretical result, does the paper provide the full set of assumptions and a complete (and correct) proof?
  \item[] Answer: \answerNA{} % Replace by \answerYes{}, \answerNo{}, or \answerNA{}.
  \item[] Justification: This paper does not include theoretical results.

  \item {\bf Experimental Result Reproducibility}
  \item[] Question: Does the paper fully disclose all the information needed to reproduce the main experimental results of the paper to the extent that it affects the main claims and/or conclusions of the paper (regardless of whether the code and data are provided or not)?
  \item[] Answer: \answerYes{}{} % Replace by \answerYes{}, \answerNo{}, or \answerNA{}.
  \item[] Justification: We have released our full datasets and baseline code alongside the paper. In addition, we have provided the details of parameter ranges used in our baseline hyper-parameter sweeps in the Appendix.

\item {\bf Open access to data and code}
  \item[] Question: Does the paper provide open access to the data and code, with sufficient instructions to faithfully reproduce the main experimental results, as described in supplemental material?
  \item[] Answer: \answerYes{} % Replace by \answerYes{}, \answerNo{}, or \answerNA{}.
  \item[] Justification: All data assets are freely shared in Wasabi with details for download on our github.io page. Our baseline code and instructions for reproducing locally are shared in GitHub.

\item {\bf Experimental Setting/Details}
  \item[] Question: Does the paper specify all the training and test details (e.g., data splits, hyperparameters, how they were chosen, type of optimizer, etc.) necessary to understand the results?
  \item[] Answer: \answerYes{}{} % Replace by \answerYes{}, \answerNo{}, or \answerNA{}.
  \item[] Justification: Dataset splits are shared as part of the data assets, and hyperparameters for the baselines are shared in the appendix. Any baseline details omitted for brevity may be retrieved from our public codebase.

\item {\bf Experiment Statistical Significance}
  \item[] Question: Does the paper report error bars suitably and correctly defined or other appropriate information about the statistical significance of the experiments?
  \item[] Answer: \answerNo{} % Replace by \answerYes{}, \answerNo{}, or \answerNA{}.
  \item[] Justification: Error bars were not provided as we believe that for the purpose of a strong baseline, an appropriately-sized hyperparameter sweep is a better use of computational resources than error bars. Code is provided so that, with minimal modification, readers may run their own baselines with error bars if required.

\item {\bf Experiments Compute Resources}
  \item[] Question: For each experiment, does the paper provide sufficient information on the computer resources (type of compute workers, memory, time of execution) needed to reproduce the experiments?
  \item[] Answer: \answerYes{} % Replace by \answerYes{}, \answerNo{}, or \answerNA{}.
  \item[] Justification: We provide details of our baselines' aggregate time and CO2 equivalent emissions. Additionally, we indicate what hardware was used to run the baselines.
  
\item {\bf Code Of Ethics}
  \item[] Question: Does the research conducted in the paper conform, in every respect, with the NeurIPS Code of Ethics \url{https://neurips.cc/public/EthicsGuidelines}?
  \item[] Answer: \answerYes{}{} % Replace by \answerYes{}, \answerNo{}, or \answerNA{}.
  \item[] Justification: Our annotators are US Employees paid at least 3x federal minimum wage. All data used was used under appropriate licenses, as discussed in the supplementary material.

\item {\bf Broader Impacts}
  \item[] Question: Does the paper discuss both potential positive societal impacts and negative societal impacts of the work performed?
  \item[] Answer: \answerYes{} % Replace by \answerYes{}, \answerNo{}, or \answerNA{}.
  \item[] Justification: The paper clearly states the intended positive impacts to research into KIE models. We briefly discuss potential for misuse in the Appendix.
  
\item {\bf Safeguards}
  \item[] Question: Does the paper describe safeguards that have been put in place for responsible release of data or models that have a high risk for misuse (e.g., pretrained language models, image generators, or scraped datasets)?
  \item[] Answer: \answerNA{} % Replace by \answerYes{}, \answerNo{}, or \answerNA{}.
  \item[] Justification: We briefly discuss possibilities for misuse in our supplementary material. However, the authors do not believe the risk is substantial enough to warrant safeguards.

\item {\bf Licenses for existing assets}
  \item[] Question: Are the creators or original owners of assets (e.g., code, data, models), used in the paper, properly credited and are the license and terms of use explicitly mentioned and properly respected?
  \item[] Answer: \answerYes{}{} % Replace by \answerYes{}, \answerNo{}, or \answerNA{}.
  \item[] Justification: In the body of the paper we have included citations for each of the document sources. Our assets are released under CC-BY-NC and the licenses or copyright information for each of the document sources are shared in the supplementary material.

\item {\bf New Assets}
  \item[] Question: Are new assets introduced in the paper well documented and is the documentation provided alongside the assets?
  \item[] Answer: \answerYes{}{} % Replace by \answerYes{}, \answerNo{}, or \answerNA{}.
  \item[] Justification: Our dataset is well documented by this paper and the associated appendices. However, at time of writing, the baseline code is only minimally documented.

\item {\bf Crowdsourcing and Research with Human Subjects}
  \item[] Question: For crowdsourcing experiments and research with human subjects, does the paper include the full text of instructions given to participants and screenshots, if applicable, as well as details about compensation (if any)? 
  \item[] Answer: \answerNo{} % Replace by \answerYes{}, \answerNo{}, or \answerNA{}.
  \item[] Justification: We have not released our labeling guides. The primary reason for this is that we used an internal labeling team who are very familiar with annotating this kind of task. As a result, the labeling guides include significant amounts of short-hand and references to internal patterns. Rather than applying post-hoc modifications to these guides to prepare for external viewing, we thought it was more appropriate to omit them entirely.

\item {\bf Institutional Review Board (IRB) Approvals or Equivalent for Research with Human Subjects}
  \item[] Question: Does the paper describe potential risks incurred by study participants, whether such risks were disclosed to the subjects, and whether Institutional Review Board (IRB) approvals (or an equivalent approval/review based on the requirements of your country or institution) were obtained?
  \item[] Answer: \answerNA{}{} % Replace by \answerYes{}, \answerNo{}, or \answerNA{}.
  \item[] Justification: We used in-house professional annotators, their work on this project was in the normal course of their employment.

\end{enumerate}

\clearpage
\section{Description of Annotation Task} \label{sec:annotation-appendix}

In this section we describe the annotation process for our datasets. For additional insights into text annotation best practices, see \citet{stollenwerk2023text}.

Prior to annotation, a set of slides was created to detail annotation expectations. Each label was allocated 1-2 slides to describe the label's intent, provide a few positive examples and document counter-examples that annotators should avoid labeling. During the annotation process, these were amended as and when clarifications were required. It is important to note that in an industry setting, the time spent by document experts annotating documents is expensive. As such, each document is seen by only one annotator, and helpful metrics like inner-annotator agreement are not available. We are mimicking this setting in the process described below.

\subsection{Annotation Interface}
A commercial annotation interface was used for all phases of annotation \cite{IndicoData_2023}. The annotation interface provides a PDF-like UI for users to apply labels via a highlighting tool, which is crucial for tasks where spatial information is necessary for accurate annotation. This approach removes any ambiguities that may have been introduced by OCR, including issues related to recognition or reading order.

In the case that the text of interest was not detected during the OCR phase, the label is necessarily omitted. This may have implications for modeling these datasets using OCR-Free approaches such as DocParser \cite{dhouib2023docparser} or Donut \cite{kim2022ocrfree}, and may make fair comparison difficult for approaches that opt to re-OCR pages using a different OCR provider. 

\subsection{Annotation Process}
The annotation process consisted of three main phases: initial annotation, model-assisted annotation, and quality review.

\subsection*{Phase 1: Initial Annotation}
Initially, the same person who developed the labeling guide annotated between 5 and 10 documents. This approach's goal is to test the labeling guide and allow for fine-tuning the schema before a wider team of professional annotators is involved. For the first 50 documents, annotation is done manually using the labeling guides and initial documents as references.

\subsection*{Phase 2: Model Assisted Annotation}
After the first 50 documents, a token-classification model is automatically trained \cite{liu2019roberta}. Predictions for this model are shown in the annotation interface, with the option to accept or reject the predictions individually or simply turn them off if they are not yet useful. The model was retrained from scratch every 50 documents, and updated predictions were shown to the annotator when available.

\subsection*{Phase 3: Quality Review}
\label{model_qa_section}
Up to this point, all documents have seen a single pass by a single annotator. A model-assisted approach was used for dataset quality assurance. After dropping all chunks that contained no labeled spans, we trained a token classification model on the dataset. We used this model to produce a spreadsheet containing all instances of disagreement between the annotations and the model predictions. We found this approach to provide a high-recall indicator of missed labels, which was the dominant error mode for long and complicated documents. For each of the datasets, a single-pass of manual review was completed using the model-label discrepancies as guidance.

\section{Additional Tables}
\label{sec:tables-appendix}

\begin{table*}[]
\centering
\begin{tabular}{l|l|l|l|l|l|l|l}
\hline
    & & Min & Max & Mean & Min & Max & Mean \\ \hline
Dataset & Num Docs & \multicolumn{3}{c|}{Num Pages}                        & \multicolumn{3}{c}{Num Words}                        \\
\hline
FCC Invoices        & 370             & 1            & 63           & 5           & 101           & 38899          & 2115         \\
S1              &13079 & 1 & 1 & 1 & 6 & 1804 & 660 \\
Resource Contracts     & 198             & 4            & 198           & 85          & 720           & 79721          & 28297        \\
NDA             & 439             & 1            & 23           & 6           & 249           & 11235          & 2705         \\
Charities          & 538             & 1            & 135           & 16          & 69           & 27308          & 3828         \\ \hline
\end{tabular}
\caption{Document length statistics for each of the datasets. Note that S1 documents have been split at the page level as discussed in Section \ref{sec:processing}. \label{tab:dataset-lengths}}
\end{table*}

\begin{table*}[htbp]
\setlength{\tabcolsep}{2pt}
\centering
\small{\begin{tabular}{|p{5,0cm}|l|l|l|l|}
\hline
Entity Name & Total Count & Unique Values & Docs Contained & Mean Entities / Doc \\
\hline
\multicolumn{5}{|c|}{\emph{fcc\_invoices}} \\
\hline
Advertiser & 1011 & 462 & 0.95 & 2.73 \\
Agency & 672 & 219 & 0.76 & 1.82 \\
Agency Commission & 373 & 291 & 0.79 & 1.01 \\
Gross Total & 818 & 376 & 0.93 & 2.21 \\
Line Item - Days & 13190 & 707 & 0.83 & 35.65 \\
Line Item - Description & 16804 & 3883 & 0.84 & 45.42 \\
Line Item - End Date & 9229 & 639 & 0.67 & 24.94 \\
Line Item - Rate & 20057 & 1134 & 0.91 & 54.21 \\
Line Item - Start Date & 19437 & 986 & 0.91 & 52.53 \\
Net Amount Due & 610 & 355 & 0.92 & 1.65 \\
Payment Terms & 439 & 32 & 0.68 & 1.19 \\
\hline
\multicolumn{5}{|c|}{\emph{s1}} \\
\hline
(Header) Description of Securities & 367 & 41 & 0.02 & 0.03 \\
(Header) Dividend Policy & 304 & 6 & 0.02 & 0.02 \\
(Header) Prospectus Summary & 320 & 6 & 0.02 & 0.02 \\
(Header) Risks To The Business & 332 & 15 & 0.03 & 0.03 \\
Agent Address & 320 & 290 & 0.02 & 0.02 \\
Agent Name & 323 & 304 & 0.02 & 0.02 \\
Agent Telephone & 311 & 279 & 0.02 & 0.02 \\
Amount Registered & 875 & 325 & 0.03 & 0.07 \\
Attorney Names & 1230 & 702 & 0.02 & 0.09 \\
Company Address & 322 & 298 & 0.02 & 0.02 \\
Company Name & 328 & 324 & 0.02 & 0.03 \\
Company Officer & 2485 & 2294 & 0.03 & 0.19 \\
Company Officer Title & 2506 & 624 & 0.03 & 0.19 \\
Date of Prospectus & 316 & 73 & 0.02 & 0.02 \\
Description of Securities (1st Para) & 374 & 337 & 0.03 & 0.03 \\
Dividend Policy (1st Para) & 307 & 267 & 0.02 & 0.02 \\
EIN & 317 & 291 & 0.02 & 0.02 \\
Joint Book Runners & 611 & 171 & 0.02 & 0.05 \\
Law Firm Address & 877 & 443 & 0.02 & 0.07 \\
Law Firm Name & 638 & 146 & 0.02 & 0.05 \\
Max Price & 493 & 164 & 0.03 & 0.04 \\
Prospectus Summary (1st Para) & 3051 & 2062 & 0.03 & 0.23 \\
Risk Clauses & 23916 & 12749 & 0.81 & 1.83 \\
Title of Security Registered & 910 & 399 & 0.03 & 0.07 \\
\hline
\multicolumn{5}{|c|}{\emph{nda}} \\
\hline
Effective Date & 420 & 373 & 0.84 & 0.96 \\
Jurisdiction & 431 & 50 & 0.95 & 0.98 \\
Party & 948 & 866 & 1.00 & 2.16 \\
\bottomrule
\end{tabular}
}
\caption{Dataset Label Frequency Statistics \label{tab:entities-1}}
\end{table*}
\begin{table*}[htbp]
\setlength{\tabcolsep}{2pt}
\centering
\small{\begin{tabular}{|p{5,0cm}|l|l|l|l|}
\hline
Entity Name & Total Count & Unique Values & Docs Contained & Mean Entities / Doc \\
\hline
\multicolumn{5}{|c|}{\emph{resource\_contracts}} \\
\hline
(Header) Contract Area Description & 252 & 152 & 0.81 & 1.27 \\
(Header) Environmental protections & 217 & 95 & 0.70 & 1.10 \\
(Header) Governing law & 233 & 58 & 0.86 & 1.18 \\
(Header) Hardship clause or force majeure & 200 & 31 & 0.88 & 1.01 \\
(Header) Income tax: rate & 177 & 62 & 0.73 & 0.89 \\
(Header) Reporting requirements & 517 & 211 & 0.82 & 2.61 \\
(Header) Term & 282 & 104 & 0.87 & 1.42 \\
(Header) Water use & 38 & 15 & 0.18 & 0.19 \\
Contract Area Description & 780 & 750 & 0.76 & 3.94 \\
Country & 432 & 118 & 0.94 & 2.18 \\
Date Signed & 290 & 219 & 0.88 & 1.46 \\
Environmental protections & 495 & 413 & 0.83 & 2.50 \\
Governing law & 246 & 180 & 0.90 & 1.24 \\
Hardship clause or force majeure & 260 & 234 & 0.91 & 1.31 \\
Income tax: rate & 186 & 166 & 0.73 & 0.94 \\
Participants & 957 & 537 & 0.95 & 4.83 \\
Project & 218 & 190 & 0.80 & 1.10 \\
Renewal or extension of term & 459 & 336 & 0.78 & 2.32 \\
Reporting requirements & 1331 & 1002 & 0.90 & 6.72 \\
Signatories, company & 568 & 399 & 0.68 & 2.87 \\
Term & 381 & 311 & 0.81 & 1.92 \\
Type & 356 & 112 & 0.99 & 1.80 \\
Water use & 130 & 84 & 0.41 & 0.66 \\
\hline
\multicolumn{5}{|c|}{\emph{charities}} \\
\hline
Accounting Basis & 373 & 336 & 0.63 & 0.69 \\
Bank Name & 368 & 138 & 0.54 & 0.68 \\
Cash In Hand at Current Year End & 482 & 418 & 0.74 & 0.90 \\
Cash In Hand at Previous Year End & 463 & 400 & 0.70 & 0.86 \\
Charity Name & 7287 & 1241 & 0.99 & 13.54 \\
Charity Registered Number & 1156 & 504 & 0.89 & 2.15 \\
Company Number & 340 & 104 & 0.17 & 0.63 \\
Event Name & 268 & 233 & 0.08 & 0.50 \\
Examination Date & 371 & 337 & 0.65 & 0.69 \\
Independent Examiner City & 913 & 425 & 0.83 & 1.70 \\
Independent Examiner Company & 768 & 397 & 0.64 & 1.43 \\
Independent Examiner Name & 660 & 519 & 0.77 & 1.23 \\
Independent Examiner Postal Code & 846 & 462 & 0.80 & 1.57 \\
Independent Examiner Street Address & 879 & 511 & 0.82 & 1.63 \\
Named Donor & 547 & 536 & 0.04 & 1.02 \\
Named Employee & 121 & 120 & 0.03 & 0.22 \\
Net Assets at Current Year End & 414 & 384 & 0.67 & 0.77 \\
Net Assets at Previous Year End & 394 & 370 & 0.65 & 0.73 \\
Net Income at Current Year End & 276 & 242 & 0.42 & 0.51 \\
Net Income at Previous Year End & 281 & 233 & 0.41 & 0.52 \\
Objectives and Activities & 477 & 465 & 0.78 & 0.89 \\
Principal Office City & 515 & 386 & 0.79 & 0.96 \\
Principal Office Postal Code & 493 & 424 & 0.77 & 0.92 \\
Principal Office Street Address & 512 & 473 & 0.80 & 0.95 \\
Project Name & 174 & 162 & 0.08 & 0.32 \\
Trustee Name & 5821 & 4652 & 0.90 & 10.82 \\
Trustee Title & 1813 & 388 & 0.74 & 3.37 \\
Year Ended & 6354 & 492 & 0.98 & 11.81 \\
\hline
\end{tabular}
}
\caption{Dataset Label Frequency Statistics Continued \label{tab:entities-1}}
\end{table*}

\begin{table*}[]
\centering
\small{\begin{tabular}{p{0.2\linewidth}|l|p{0.2\linewidth}}
Parameter          & Distribution & Value Range \\
\hline
max empty chunk ratio    & log uniform & {[}1e-2, 1000{]}\\
\hline
learning rate        & log uniform & {[}1e-8, 1e-2{]}\\
\hline
batch\_size         & uniform   & {[}1, max{]}\\
\hline
num epochs         & uniform   &{[}1, 16/64{]}\\
\hline
learning rate warmup    & uniform   & {[}0, 0.5{]}\\
\hline
collapse whitespace     &       & {[}True, False{]}\\
\hline
max gradient norm      & log uniform & {[}1e-3, 1e5{]}\\
\hline
L2 regularization      & log uniform & {[}1e-5, 1.0{]}\\
\hline
gradient accumulation steps & uniform   & {[}1, 8{]}\\
\hline
learning rate schedule   &       & {[}linear, cosine, cosine\_with\_restarts, constant, constant\_with\_warmup, inverse\_sqrt{]}
\end{tabular}
}
\caption{Sweep parameters and ranges for baselines. Note that for LayoutLM-v3, we observed that 16 epochs were insufficient to produce a viable baseline, so we extended this to 64 epochs; all other models used a maximum of 16 epochs.}
\label{tab:sweep_parameters}
\end{table*}

\begin{table*}
\begin{adjustwidth}{-1in}{-1in}
\label{tab/field_desc_fcc_invoices}
\begin{tabular}{p{3cm}|p{3cm}|p{4cm}|p{7cm}}
\toprule
Field Name & Aliases & Description & Notes \\ \midrule
\midrule
Advertiser & 'Advertiser', 'Client Name' & Legal, trade, or DBA name of the advertiser being represented. & Can be shown as 'Bill To' or 'Customer'. Will be found in every document. Do not capture name after 'Attn:' \\ \midrule
Agency & 'Bill To' & Business managing the advertising captaign for client/ advertiser & \\ \midrule
Agency Commission & 'Commissions' & Dollar value of the commission obtained by Agency. & Found at the bottom of invoice under the 'Totals' segment. Can be called out as 'Commission.' \\ \midrule
Gross Total & 'Gross Billings' & Total amount of advertisement billed prior to deducted agency commission. & Found at the bottom of invoice under 'Totals' segment. Will have explicit callout. \\ \midrule
Line Item - Days & 'Buy Line Dates', 'Air Time', 'Weekday', 'MTWTFSS' & Day of which advertisement is scheduled to air. & Can be found under indepenedently separated column or at the end of a combined column. Must be captured within line items.  \\ \midrule
Line Item - Description & 'Program', 'Description' & Name of program of which advertisement will air within.  & Capture under 'Description' if clearly called out. Can also be shown as 'Program'. Must be captured within line items.  \\ \midrule
Line Item - End Date &  & Final date of advertisement aired run. & Will not always be explicitly called out as 'End Date.' \\ \midrule
Line Item - Rate & 'Amount', 'Rate' & Single advertisement airing cost. & Will not always have an explicit callout, but will always be found within advertisement line items. \\ \midrule
Line Item - Start Date &  & First date of advertisement aired run.  & Will not always be explictly called out as 'Start Date.' \\ \midrule
Net Amount Due &  & Total amount of advertisement billed after Agency Commission is deducted from Gross Total.  & Found at the bottom of invoice under the 'Totals' segment. \\ \midrule
Payment Terms & 'Terms' & Explicit definition as to how advertisement cost is to be paid for.  & Occassionally will be hidden within the fine print. Can show up as 'Cash In Advance', 'Net 30'.  \\ \midrule
\bottomrule
\end{tabular}
\caption{Field Descriptions for FCC Invoices}
\end{adjustwidth}
\end{table*}

\begin{table*}
\begin{adjustwidth}{-1in}{-1in}
\label{tab/field_desc_charity_reports_v7}
\begin{tabular}{p{3cm}|p{3cm}|p{4cm}|p{7cm}}
\toprule
Field Name & Aliases & Description & Notes \\ \midrule
\midrule
Accounting Basis & 'Accounting Policies', 'Responsibilities and Basis of Report' & Laws and regulation of which the financial statements are prepared by. & Can be found under 'Notes to the Financial Statement'. Capture entire segment/clause. \\ \midrule
Bank Name & 'Bank', 'Banker', 'Bankers' & Name of financial institution the charity holds their funds within.  & Explicitly called out. Typically will be found under recitals to the charity. Include 'PLC' or similar included in institution name.  \\ \midrule
Cash In Hand at Current Year End & 'Cash In Hand', 'Cash at bank and in hand' & Amount of funds held within charity. & Found within financials or Balance Sheet of charity report. Will be listed under 'Current Assets'. Capture amount under current reported year. \\ \midrule
Cash In Hand at Previous Year End & 'Cash In Hand', 'Cash at bank and in hand' & Amount of funds held within charity. & Found within financials or Balance Sheet of charity report. Will be listed under 'Current Assets'. Capture amount under previously reported year. \\ \midrule
Charity Name &  & Name of the Charity.  & Explicitly called out throughout the entirety of the document. Can typically be found within headers and/ or footers. \\ \midrule
Charity Registered Number &  & Identifying registration number identifying charity as a charitable organization.  & Explicitly called out throughout the entirety of the document. Can typically be found within the first 5-10 pages of document consistently, including cover page.  \\ \midrule
Company Number &  & Companies House registration number. Will be 8 digits or 2 letters followed by 6 digits. & Explicitly called out within the document. Can typically be found within the first 5-10 pages of document consistently, including cover page. Capture all instances, including cover page.  \\ \midrule
Event Name &  & Single occurrence gatherings of which charity funds are used & Varying location of event names. Can be found within 'Activities of the Charity'. Capture all event names. Will be a single event of which does not occur more than once. Will need to read through event description to determine of Event or Project.  \\ \midrule
Examination Date &  & Date on which charity report was examined by independent examiner.  & Can be found on 'Independent Examiner's Report to the Trustees' segment. Located near additional instance of independent examiner details. Can be found near examiner signature.  \\ \midrule
Independent Examiner City &  & City and County name where the independent examiner company is based.  & Can be found on 'Independent Examiner's Report to the Trustees' segment.  Can be found near examiner signature. \\ \midrule
Independent Examiner Company &  & Company name of which Independent examiner is representing.  & Can be found on 'Independent Examiner's Report to the Trustees' segment.  Can be found near examiner signature. \\ \midrule
Independent Examiner Name &  & Name of independent examiner/ individual compiling charity report. & Can be found on 'Independent Examiner's Report to the Trustees' segment.  Can be found near examiner signature. Do not capture titles such as 'Chartered Accountant', 'ACA', etc. \\ \midrule
Independent Examiner Postal Code &  & Zip Code/ Postal Code of independent examiner company. & Can be found on 'Independent Examiner's Report to the Trustees' segment.  Can be found near examiner signature.  \\ \midrule
Independent Examiner Street Address &  & Street Address and building name [if applicable] of independent examiner company. & Can be found on 'Independent Examiner's Report to the Trustees' segment.  Can be found near examiner signature.  \\ \midrule
\bottomrule
\end{tabular}
\caption{Field Descriptions for Charities}
\end{adjustwidth}
\end{table*}

\begin{table*}
\begin{adjustwidth}{-1in}{-1in}
\label{tab/field_desc_charity_reports_v7}
\begin{tabular}{p{3cm}|p{3cm}|p{4cm}|p{7cm}}
\toprule
Field Name & Aliases & Description & Notes \\ \midrule
\midrule
Named Donor &  & Individual, organization, or company name of entity of which financially donated to charity.  & Can be found under 'List of Principal Donors'. Include punctuation such as dashes within donor name.  \\ \midrule
Named Employee & 'Senior Leadership Team', 'Key Management Personnel' & Name of hired staff to manage the charity. & Include titles such as Mr, Mrs, etc.  \\ \midrule
Net Assets at Current Year End &  & Total value of charity's assets minus its liabilities for current reported year. & Found in the financials or 'Balance Sheet' of charity report. Utilize dates shown in financials to determine currently reported year.  \\ \midrule
Net Assets at Previous Year End &  & Total value of charity's assets minus it's liabilities for the previously reported year.  & Found in the financials or 'Balance Sheet' of charity report. Utilize dates shown in financials to determine previously reported year.  \\ \midrule
Net Income at Current Year End &  & Total value of income after deductions of expenses, taxes, etc for the current reported year.  & Found in the financials or 'Balance Sheet' of charity report. Utilize listed above dates to determine currently reported year. \\ \midrule
Net Income at Previous Year End &  & Total value of income after deductions of expenses, taxes, etc for the previously reported year.  & Found in the financials or 'Balance Sheet' of charity report. Utilize listed above dates to determine previously reported year. \\ \midrule
Objectives and Activities & 'Objectives and aims' & Goal, intention, and actions of the charity.  & Found within the first 5-10 pages of report and will be clearly called out. \\ \midrule
Principal Office City & 'Registered Office' & City and county of charity's principal location.  & Found within first 5-10 pages of document.  \\ \midrule
Principal Office Postal Code & 'Registered Office' & Zip Code/ Postal Code of charity's principal location.  & Found within first 5-10 pages of document. \\ \midrule
Principal Office Street Address & 'Registered Office' & Street Address and building name [if applicable] of charity's principal location.  & Found within first 5-10 pages of document.  \\ \midrule
Project Name &  & Reoccuring programs of which charity funds are used. & Often found within 'Activities of the Charity'. Capture all project names. Will be an event that happens more than once. Will need to read through event description to determine if Event or Project.  \\ \midrule
Trustee Name & 'Trustees', 'Govenors', 'Elders', 'Deacons' & Volunteers who lead the charity and decide how they operate.  & Found within first 5-10 pages of document. Capture single instance.  \\ \midrule
Trustee Title & 'Chairman', 'Chair', 'Administrator', 'Vice-Chairman', 'Treasurer' & Volunteer titles of those whom lead the charity & Found within first 5-10 pages of document. \\ \midrule
Year Ended &  & The end date of which the report is representative of.  &  \\ \midrule
\bottomrule
\end{tabular}
\caption{Field Descriptions for Charities}
\end{adjustwidth}
\end{table*}

\begin{table*}
\begin{adjustwidth}{-1in}{-1in}
\label{tab/field_desc_resource_contracts}
\begin{tabular}{p{3cm}|p{3cm}|p{4cm}|p{7cm}}
\toprule
Field Name & Aliases & Description & Notes \\ \midrule
\midrule
(Header) Contract Area Description & 'Contract Area' & Section title to the Contract Area Description & Capture clause/ segment number with section header. \\ \midrule
(Header) Environmental protections &  & Section title to the declaration of environmental and safety measures, as well as the prevention of loss. & Header may not always be present to identify clause. Will sometimes show near other safety measures. Capture clause/ segment number with section header. \\ \midrule
(Header) Governing law &  & Section title to the set of laws used to enforce stipulations of contract.  & Header typically will always occur prior to clause. Capture clause/ segment number with section header. \\ \midrule
(Header) Hardship clause or force majeure &  & Section title to the clause indicating the stipulations of an unforeseen event occurring during the course of a contract. & Header will always typically occur prior to clause. Capture clause/ segment number with section header. \\ \midrule
(Header) Income tax: rate & 'Income Tax and Allowable Contract Expenditures' & Section title to the tax laws and/ or tax code of a specific/ involved country.  & Header will always typically occur prior to clause. Capture clause/ segment number with section header. \\ \midrule
(Header) Reporting requirements & 'General responsibilities of the contractor', 'Accounting procedures' & Section title to the reporting stipulations of the agreement. & Can show with a variety of headers. Capture clause/ segment number with section header. \\ \midrule
(Header) Term & 'Effective Date and Term', 'Contract Effective Period' & Section title to the period of which the agreement is effective. & Can show with a variety of headers. Capture clause/ segment number with section header. \\ \midrule
(Header) Water use &  & Section title to the usage, rights to, and purpose of water pertaining to the agreement. & Does not always show with a header, but can be found within the environmental clauses. Capture segment number with section header. \\ \midrule
Contract Area Description & 'Contract Area' & Description of the location of which the agreement has been made for.  & Capture entire clause/ segment, including area coordinates and any coordinate tables within the section.  \\ \midrule
Country &  & Name of country executing the contract.  & Capture in the pages before the definitions only, and capture all instances within this range. May occur several times in this section before the definitions.  \\ \midrule
Date Signed &  & Date of which contract is signed. & Can show up in many places, but focus on capturing from the first few pages before the definitions and adjacent to the signatories.  \\ \midrule
Environmental protections &  & Declaration of environmental and safety measures, as well as the prevention of loss. & Header may not always be present to identify clause. Will sometimes show near other safety measures. Only captured clauses that explicitly refer to environmental protection or other natural resources and parks. Capture entire clause/ statement.  \\ \midrule
\bottomrule
\end{tabular}
\caption{Field Descriptions for Resource Contracts}
\end{adjustwidth}
\end{table*}

\begin{table*}
\begin{adjustwidth}{-1in}{-1in}
\label{tab/field_desc_resource_contracts}
\begin{tabular}{p{3cm}|p{3cm}|p{4cm}|p{7cm}}
\toprule
Field Name & Aliases & Description & Notes \\ \midrule
\midrule
Governing law &  & Set of laws used to enforce stipulations of contract.  & Capture entire clause/ segment. Often a related clause about which laws are used in cases of arbitration - do not capture those clauses. Typically will have a header indicating the Governing Law clause.  \\ \midrule
Hardship clause or force majeure &  & Clause indicating the stipulations of an unforeseen event occuring during the course of a contract. & Capture segment of clause that defines what a force majeure is. Will rarely show up within the definitions segment of the document.  \\ \midrule
Income tax: rate & 'Income Tax and Allowable Contract Expenditures' & Tax laws and/ or tax code of a specific/ involved country.  & Capture specific clause indicating government taxes on income/ profits. Typically specifies the tax laws of a given country or a more specific tax code.  \\ \midrule
Participants &  & Names of corporations and sovereign entities participating in the contract. & Will occur alongside country and contract type. This callout typically shows within the preamble/ recitals of the contract.  \\ \midrule
Project &  & Name of the region where mining/ exploration listed within the agreement occurs.  & Typically occurs along other header fields, prior to the definitions section.  \\ \midrule
Renewal or extension of term &  & Clause indicating the periods of which the contract can be renewed or extended.  & Capture explicit statement indicating the terms of contract extension and renewal terms.  \\ \midrule
Reporting requirements & 'General responsibilities of the contractor', 'Accounting procedures' & Reporting stipulations of the agreement. & Can show up in varied locations throughout a single document.  Capture entire clause description the stipulations of the reporting requirements of the agreement.  \\ \midrule
Signatories, company &  & Names of the individuals representing the government or participating companies to the agreement. & Can occur in variety of places throughout the document, but generally towards the end. Signatures can occasionally be handwritten. \\ \midrule
Term & 'Effective Date and Term', 'Contract Effective Period' & Period of which the agreement is effective. & Capture section of clause pertaining to the overall agreement term, including the exploration, production terms, but excluding renewal and extension terms.  \\ \midrule
Type &  & Agreement identifier as to which kind of agreement this document is curated for.  & Typically occurs within the first few pages prior to the definitions segment.  \\ \midrule
Water use &  & Usage, rights to, and purpose of water pertaining to the agreement. & Does not always show within a header, but can be found within the environmental clauses. Capture entire clause pertaining to the agreed usage of water within the agreement.  \\ \midrule
\bottomrule
\end{tabular}
\caption{Field Descriptions for Resource Contracts}
\end{adjustwidth}
\end{table*}

\begin{table*}
\begin{adjustwidth}{-1in}{-1in}
\label{tab/field_desc_s1}
\begin{tabular}{p{3cm}|p{3cm}|p{4cm}|p{7cm}}
\toprule
Field Name & Aliases & Description & Notes \\ \midrule
\midrule
(Header) Description of Securities &  & Section header to the description of all registered securities.  & Header typically clearly called out. Capture clause/ segment number with header.  \\ \midrule
(Header) Dividend Policy &  & Section header to the company plan for earnings distribution to shareholders. & Header typically clearly called out. Capture clause/ segment number with header.  \\ \midrule
(Header) Prospectus Summary &  & Section header to the summarized description of a company, it's services, goals, and plans before or during the sale of a company. & Header occasionally will show as vague as 'Summary'. Capture clause/ segment number with header.  \\ \midrule
(Header) Risks To The Business & 'Risk Factors' & Section header to the outlying risks of a business.  & Header typically clearly called out. Capture clause/ segment number with header.  \\ \midrule
Agent Address &  & Business address of agent involved in the sale of a company.  & Typically shown within the first few pages of the document within the recitals. Will usually be found in the middle section of the page, and occasionally will be clearly called out. Include street, suite, city, state, and zip code in captures. \\ \midrule
Agent Name &  & Name of agent or business involved in the sale of a company. & Typically shown within the first few pages of the document within the recitals. Will usually be found in the middle section of the page, and occasionally will be clearly called out. Include parenthesized state and 'Inc.' within the capture.  \\ \midrule
Agent Telephone &  & Phone number of agent involved in the sale of a company. & Typically shown within the first few pages of the document within the recitals. Will usually be found in the middle section of the page, and occasionally will be clearly called out. \\ \midrule
Amount Registered & 'Amount Being Registered' & Total amount of registered Units, Shares, or Warrants per each class of registered security. & Can be found within the calculation of registration fee. Will be clearly called out. Capture kind of unit with each amount registered.  \\ \midrule
Attorney Names &  & Names of attorneys retained/ in relation with the company.  & Typically shown within the first few pages of the document within the recitals. Will usually be found in the middle-bottom section of the page, and occasionally will be clearly called out. \\ \midrule
Company Address &  & Address of company agreement pertains to. & Typically shown within the first few pages of the document within the recitals. Will usually be found in the top section of the page, and will be clearly called out. Include street, suite, city, state, and zip code in captures. \\ \midrule
Company Name &  & Name of company agreement pertains to. & Typically shown within the first few pages of the document within the recitals. Will usually be found in the top section of the page, and will be clearly called out. Include parenthesized state and 'Inc.' within the capture.  \\ \midrule
Company Officer &  & Name of company officer. & Capture all identified officer names. \\ \midrule
\bottomrule
\end{tabular}
\caption{Field Descriptions for S1}
\end{adjustwidth}
\end{table*}

\begin{table*}
\begin{adjustwidth}{-1in}{-1in}
\label{tab/field_desc_s1}
\begin{tabular}{p{3cm}|p{3cm}|p{4cm}|p{7cm}}
\toprule
Field Name & Aliases & Description & Notes \\ \midrule
\midrule
Company Officer Title & 'Co-founder', 'Chairman', 'Chief Executive Officer', 'Director Nominee' & Title of company officer.  & Capture all identified officer titles.  \\ \midrule
Date of Prospectus &  & Date of initial public offering being published. & Explicitly called out as such. \\ \midrule
Description of Securities (1st Para) &  & Description of all registered securities.  & Detailed description of all securities registered to said agreement. Capture entire first clause/ paragraph.  \\ \midrule
Dividend Policy (1st Para) &  & Company plan for earnings distribution to shareholders. & Detailed description of how earnings are distributed among shareholders. On occasion will show as multiple paragraphs/ clauses; capture first paragraph only in this case. \\ \midrule
EIN & 'IRS Employer Identification Number', 'Employer Identification Number' & Identifier assigned to a business which is recognized by the IRS & Typically shown within the first few pages of the document within the recitals. Will usually be found in the top section of the page, and will be clearly called out. \\ \midrule
Joint Book Runners & 'Joint Book-Running Managers' & Conjoined firms responsible for the underwriting process during an Initial Public Offering [IPO] or a leveraged buyout.  & Will be explicitly called out. If multiple firms identified, capture all in specified segment of document.  \\ \midrule
Law Firm Address & 'Copies to:' & Address of law firms of which attorneys represent  & Typically shown within the first few pages of the document within the recitals. Will usually be found in the bottom section of the page, and will be clearly called out. Capture Street, City, State, Province, ZIP as a part of address. \\ \midrule
Law Firm Name & 'Copies to:' & Name of law firm of which attorneys represent & Typically shown within the first few pages of the document within the recitals. Will usually be found in the bottom section of the page, and will be clearly called out. Capture 'LLP' as a part of firm name. \\ \midrule
Max Price & 'Proposed Maximum Aggregate Offering Price' & The maximum amount incorporating the sum of all cash, services, property, notes, cancellation of debt, or other consideration to be received by an issuer for issuance of it's securities.  & Can be found within the calculation of registration fee. Will be clearly called out. Capture currency with field. \\ \midrule
Prospectus Summary (1st Para) &  & Summarized description of a company, it's services, goals, and plans before or during the sale of a company. & Summary is often broken into bullet points, but can also present as a long-form paragraph. Each bullet should be captured independently. Do not include bullet punctuation in capture.  \\ \midrule
Risk Clauses & 'Risk Factors' & Outlying risks of a business.  & Typically will show as bold and/or italicized sections within the document. Capture summarized statements such as these.  \\ \midrule
Title of Security Registered &  & Name/ description of the type of investment. & Can be found within the calculation of registration fee. Will be clearly called out. Do not capture article number with title. \\ \midrule
\bottomrule
\end{tabular}
\caption{Field Descriptions for S1}
\end{adjustwidth}
\end{table*}

\begin{table*}[h!]
\begin{adjustwidth}{-1in}{-1in}
\centering\small{
\begin{tabular}{|c|c|c|c|c|c|c|c|c|c|c|c|}
\hline
\textbf{Dataset} & \textbf{Base Model} & \textbf{F1} & \textbf{\rot{Learning Rate}} & \textbf{\rot{Num Epochs}} & \textbf{\rot{Batch Size}} & \textbf{\rot{Gradient Accumulation Steps }} & \textbf{\rot{L2 Regularization}} & \textbf{\rot{Max Empty Chunk Ratio}} & \textbf{\rot{Max Grad Norm}} & \textbf{\rot{LR Warmup}} & \textbf{\rot{LR Schedule}}\\
\hline
\hline\multirow{4}{*}{Charities} & Longformer Base & 58.1 & 7.7E-5 & 15 & 1 & 7 & 2.1E-1 & 56.63 & 1.9E-2 & 0.43 & constant\_with\_warmup \\
\cline{2-12}
 & \textbf{LayoutLM-v3 Base} & \textbf{63.6} & \textbf{1.2E-5} & \textbf{58} & \textbf{2} & \textbf{3} & \textbf{5.3E-4} & \textbf{0.31} & \textbf{5.3E-3} & \textbf{0.46} & \textbf{constant} \\
\cline{2-12}
 & DeBERTa-v3 Base & 61.3 & 1.1E-5 & 16 & 1 & 7 & 3.3E-4 & 0.66 & 1.4E-3 & 0.00 & constant \\
\cline{2-12}
 & RoBERTa Base & 57.6 & 3.9E-5 & 15 & 3 & 4 & 3.2E-4 & 58.57 & 1.7E+04 & 0.13 & constant\_with\_warmup \\
\hline\multirow{4}{*}{FCC Invoices} & Longformer Base & 67.3 & 1.2E-4 & 11 & 1 & 1 & 3.5E-1 & 9.43 & 6.5E-2 & 0.20 & cosine \\
\cline{2-12}
 & LayoutLM-v3 Base & 68.3 & 9.1E-6 & 33 & 1 & 1 & 9.0E-3 & 124.26 & 8.5E-2 & 0.09 & constant \\
\cline{2-12}
 & \textbf{DeBERTa-v3 Base} & \textbf{69.2} & \textbf{7.2E-5} & \textbf{12} & \textbf{1} & \textbf{1} & \textbf{9.0E-2} & \textbf{0.16} & \textbf{5.2E-3} & \textbf{0.39} & \textbf{cosine\_with\_restarts} \\
\cline{2-12}
 & RoBERTa Base & 66.5 & 5.0E-5 & 10 & 1 & 3 & 8.0E-4 & 0.11 & 2.1E+04 & 0.04 & inverse\_sqrt \\
\hline\multirow{4}{*}{NDA} & Longformer Base & 81.0 & 7.8E-5 & 12 & 1 & 2 & 2.1E-2 & 489.44 & 5.3E-3 & 0.34 & linear \\
\cline{2-12}
 & LayoutLM-v3 Base & 80.7 & 7.6E-6 & 25 & 1 & 8 & 6.7E-4 & 5.58 & 6.7E-1 & 0.09 & constant\_with\_warmup \\
\cline{2-12}
 & \textbf{DeBERTa-v3 Base} & \textbf{83.7} & \textbf{8.1E-5} & \textbf{9} & \textbf{1} & \textbf{8} & \textbf{3.1E-4} & \textbf{40.67} & \textbf{5.4E-3} & \textbf{0.46} & \textbf{constant} \\
\cline{2-12}
 & RoBERTa Base & 81.5 & 1.0E-5 & 9 & 2 & 3 & 4.9E-1 & 37.47 & 1.2E-1 & 0.15 & cosine\_with\_restarts \\
\hline\multirow{4}{*}{Resource Contracts} & Longformer Base & 45.5 & 2.6E-5 & 14 & 1 & 1 & 2.6E-2 & 47.14 & 2.2E-2 & 0.15 & cosine\_with\_restarts \\
\cline{2-12}
 & LayoutLM-v3 Base & 41.8 & 8.2E-6 & 14 & 2 & 6 & 3.8E-4 & 8.32 & 4.6E-2 & 0.41 & cosine\_with\_restarts \\
\cline{2-12}
 & \textbf{DeBERTa-v3 Base} & \textbf{45.6} & \textbf{1.5E-5} & \textbf{16} & \textbf{2} & \textbf{3} & \textbf{1.7E-4} & \textbf{5.77} & \textbf{9.1E-3} & \textbf{0.14} & \textbf{constant\_with\_warmup} \\
\cline{2-12}
 & RoBERTa Base & 40.9 & 4.5E-5 & 11 & 3 & 4 & 1.1E-1 & 20.20 & 8.6E-3 & 0.38 & inverse\_sqrt \\
\hline\multirow{4}{*}{S1} & Longformer Base & 82.6 & 4.0E-5 & 13 & 1 & 2 & 4.4E-2 & 144.02 & 6.5E-3 & 0.44 & cosine\_with\_restarts \\
\cline{2-12}
 & \textbf{LayoutLM-v3 Base} & \textbf{83.5} & \textbf{1.8E-5} & \textbf{27} & \textbf{1} & \textbf{7} & \textbf{9.4E-4} & \textbf{0.21} & \textbf{5.3E-1} & \textbf{0.25} & \textbf{cosine\_with\_restarts} \\
\cline{2-12}
 & DeBERTa-v3 Base & 81.8 & 1.0E-5 & 13 & 1 & 1 & 3.5E-5 & 5.24 & 3.5E+00 & 0.33 & cosine\_with\_restarts \\
\cline{2-12}
 & RoBERTa Base & 81.7 & 4.1E-6 & 11 & 1 & 2 & 3.6E-1 & 0.19 & 4.0E-1 & 0.29 & constant \\
\hline
\end{tabular}}
\end{adjustwidth}
\caption{Model parameters that demonstrated the highest Validation-set F1 for each model and dataset pair. The F1 column corresponds to test-set F1. Note that batch size and gradient accumulation steps are included separately. The number of epochs for DeBERTa-v3 on Charities and Resource contracts is at the limit of the sweep range. This may indicate that our parameter range was too restrictive for this model.}
\label{tab:best_params}
\end{table*}

\begin{table*}\centering\scriptsize{\begin{tabular}{|c|c|c|c|c|}
\hline
\textbf{Field Name} & \textbf{Base Model} & \textbf{Test F1} & \textbf{Test Precision} & \textbf{Test Recall} \\
\hline
\hline
\multirow{4}{*}{Accounting Basis} & Longformer Base & 33.6 & 64.2 & 22.8 \\
\cline{2-5}
 & \textbf{LayoutLM-v3 Base} & \textbf{60.4} & \textbf{67.2} & \textbf{54.9} \\
\cline{2-5}
 & DeBERTa-v3 Base & 46.2 & 49.3 & 43.4 \\
\cline{2-5}
 & RoBERTa Base & 27.8 & 56.7 & 18.4 \\
\hline
\multirow{4}{*}{Bank Name} & Longformer Base & 76.3 & \textbf{76.9} & 75.8 \\
\cline{2-5}
 & \textbf{LayoutLM-v3 Base} & \textbf{80.0} & 70.8 & \textbf{92.0} \\
\cline{2-5}
 & DeBERTa-v3 Base & 75.4 & 70.8 & 80.7 \\
\cline{2-5}
 & RoBERTa Base & 71.1 & 73.8 & 68.6 \\
\hline
\multirow{4}{*}{Cash In Hand at Current Year End} & Longformer Base & 62.7 & 58.0 & 68.2 \\
\cline{2-5}
 & LayoutLM-v3 Base & 62.2 & \textbf{60.0} & 64.5 \\
\cline{2-5}
 & \textbf{DeBERTa-v3 Base} & \textbf{68.6} & 59.0 & \textbf{81.9} \\
\cline{2-5}
 & RoBERTa Base & 58.3 & 51.0 & 68.0 \\
\hline
\multirow{4}{*}{Cash In Hand at Previous Year End} & Longformer Base & 60.9 & \textbf{62.5} & 59.4 \\
\cline{2-5}
 & LayoutLM-v3 Base & 64.1 & 60.4 & 68.2 \\
\cline{2-5}
 & \textbf{DeBERTa-v3 Base} & \textbf{66.7} & 56.2 & \textbf{81.8} \\
\cline{2-5}
 & RoBERTa Base & 62.1 & 52.1 & 76.9 \\
\hline
\multirow{4}{*}{Charity Name} & Longformer Base & 82.4 & 83.8 & 80.9 \\
\cline{2-5}
 & LayoutLM-v3 Base & 82.4 & \textbf{84.2} & 80.7 \\
\cline{2-5}
 & DeBERTa-v3 Base & 81.8 & 83.2 & 80.5 \\
\cline{2-5}
 & \textbf{RoBERTa Base} & \textbf{82.9} & \textbf{84.2} & \textbf{81.6} \\
\hline
\multirow{4}{*}{Charity Registered Number} & Longformer Base & 91.7 & 90.8 & \textbf{92.6} \\
\cline{2-5}
 & LayoutLM-v3 Base & 90.4 & 92.4 & 88.5 \\
\cline{2-5}
 & \textbf{DeBERTa-v3 Base} & \textbf{92.2} & \textbf{95.6} & 89.1 \\
\cline{2-5}
 & RoBERTa Base & 91.6 & 94.4 & 89.0 \\
\hline
\multirow{4}{*}{Company Number} & Longformer Base & 83.0 & 93.8 & 74.4 \\
\cline{2-5}
 & LayoutLM-v3 Base & 88.6 & \textbf{95.4} & 82.7 \\
\cline{2-5}
 & \textbf{DeBERTa-v3 Base} & \textbf{91.2} & \textbf{95.4} & \textbf{87.3} \\
\cline{2-5}
 & RoBERTa Base & 79.5 & 89.2 & 71.6 \\
\hline
\multirow{4}{*}{Event Name} & Longformer Base & 15.7 & 13.8 & 18.2 \\
\cline{2-5}
 & \textbf{LayoutLM-v3 Base} & \textbf{22.7} & \textbf{34.5} & 16.9 \\
\cline{2-5}
 & DeBERTa-v3 Base & 16.5 & 13.8 & 20.5 \\
\cline{2-5}
 & RoBERTa Base & 18.9 & 15.5 & \textbf{24.3} \\
\hline
\multirow{4}{*}{Examination Date} & Longformer Base & 74.9 & 78.0 & 71.9 \\
\cline{2-5}
 & LayoutLM-v3 Base & 73.3 & 67.1 & \textbf{80.9} \\
\cline{2-5}
 & DeBERTa-v3 Base & 75.2 & 75.6 & 74.7 \\
\cline{2-5}
 & \textbf{RoBERTa Base} & \textbf{77.0} & \textbf{81.7} & 72.8 \\
\hline
\multirow{4}{*}{Independent Examiner City} & Longformer Base & 81.3 & 79.2 & 83.4 \\
\cline{2-5}
 & \textbf{LayoutLM-v3 Base} & \textbf{85.3} & \textbf{86.5} & \textbf{84.2} \\
\cline{2-5}
 & DeBERTa-v3 Base & 79.0 & 79.2 & 78.8 \\
\cline{2-5}
 & RoBERTa Base & 79.0 & 81.5 & 76.7 \\
\hline
\multirow{4}{*}{Independent Examiner Company} & \textbf{Longformer Base} & \textbf{71.9} & 75.2 & \textbf{69.0} \\
\cline{2-5}
 & LayoutLM-v3 Base & 69.4 & 71.0 & 67.8 \\
\cline{2-5}
 & DeBERTa-v3 Base & 67.1 & 70.3 & 64.2 \\
\cline{2-5}
 & RoBERTa Base & 66.9 & \textbf{75.9} & 59.8 \\
\hline
\multirow{4}{*}{Independent Examiner Name} & Longformer Base & 72.8 & 68.8 & 77.2 \\
\cline{2-5}
 & \textbf{LayoutLM-v3 Base} & \textbf{77.4} & \textbf{73.2} & \textbf{82.1} \\
\cline{2-5}
 & DeBERTa-v3 Base & 68.2 & 65.9 & 70.5 \\
\cline{2-5}
 & RoBERTa Base & 74.5 & \textbf{73.2} & 75.9 \\
\hline
\multirow{4}{*}{Independent Examiner Postal Code} & Longformer Base & 90.2 & 90.7 & 89.6 \\
\cline{2-5}
 & \textbf{LayoutLM-v3 Base} & \textbf{94.3} & \textbf{96.3} & \textbf{92.3} \\
\cline{2-5}
 & DeBERTa-v3 Base & 92.7 & 93.8 & 91.6 \\
\cline{2-5}
 & RoBERTa Base & 91.3 & 93.8 & 88.9 \\
\hline
\multirow{4}{*}{Independent Examiner Street Address} & \textbf{Longformer Base} & \textbf{82.8} & \textbf{84.2} & \textbf{81.4} \\
\cline{2-5}
 & LayoutLM-v3 Base & 81.7 & 82.5 & 81.0 \\
\cline{2-5}
 & DeBERTa-v3 Base & 80.8 & 82.5 & 79.2 \\
\cline{2-5}
 & RoBERTa Base & 80.2 & 83.0 & 77.6 \\
\hline
\multirow{4}{*}{Named Donor} & Longformer Base & 0.0 & 0.0 & 0.0 \\
\cline{2-5}
 & LayoutLM-v3 Base & 0.0 & 0.0 & 0.0 \\
\cline{2-5}
 & \textbf{DeBERTa-v3 Base} & \textbf{1.8} & \textbf{1.1} & \textbf{4.3} \\
\cline{2-5}
 & RoBERTa Base & 0.0 & 0.0 & 0.0 \\
\hline
\multirow{4}{*}{Named Employee} & Longformer Base & 0.0 & 0.0 & 0.0 \\
\cline{2-5}
 & LayoutLM-v3 Base & 0.0 & 0.0 & 0.0 \\
\cline{2-5}
 & \textbf{DeBERTa-v3 Base} & \textbf{3.3} & \textbf{1.9} & \textbf{12.5} \\
\cline{2-5}
 & RoBERTa Base & 0.0 & 0.0 & 0.0 \\
\hline
\multirow{4}{*}{Net Assets at Current Year End} & Longformer Base & 58.1 & 62.5 & 54.3 \\
\cline{2-5}
 & \textbf{LayoutLM-v3 Base} & \textbf{72.4} & \textbf{68.8} & \textbf{76.4} \\
\cline{2-5}
 & DeBERTa-v3 Base & 67.5 & 63.7 & 71.8 \\
\cline{2-5}
 & RoBERTa Base & 59.9 & 62.5 & 57.5 \\
\hline
\multirow{4}{*}{Net Assets at Previous Year End} & Longformer Base & 59.0 & 59.7 & 58.2 \\
\cline{2-5}
 & \textbf{LayoutLM-v3 Base} & \textbf{72.0} & \textbf{70.1} & \textbf{74.0} \\
\cline{2-5}
 & DeBERTa-v3 Base & 66.7 & 63.6 & 70.0 \\
\cline{2-5}
 & RoBERTa Base & 59.4 & 67.5 & 53.1 \\
\hline
\end{tabular}}\caption{Field level metrics for Charities.}\end{table*}

\begin{table*}\centering\scriptsize{\begin{tabular}{|c|c|c|c|c|}
\hline
\textbf{Field Name} & \textbf{Base Model} & \textbf{Test F1} & \textbf{Test Precision} & \textbf{Test Recall} \\
\hline
\hline
\multirow{4}{*}{Net Income at Current Year End} & Longformer Base & 30.5 & 28.6 & 32.7 \\
\cline{2-5}
 & LayoutLM-v3 Base & 48.6 & \textbf{48.2} & 49.1 \\
\cline{2-5}
 & \textbf{DeBERTa-v3 Base} & \textbf{50.5} & 46.4 & \textbf{55.3} \\
\cline{2-5}
 & RoBERTa Base & 19.6 & 16.1 & 25.0 \\
\hline
\multirow{4}{*}{Net Income at Previous Year End} & Longformer Base & 50.0 & 49.2 & 50.9 \\
\cline{2-5}
 & \textbf{LayoutLM-v3 Base} & \textbf{54.7} & \textbf{54.2} & \textbf{55.2} \\
\cline{2-5}
 & DeBERTa-v3 Base & 49.6 & 47.5 & 51.9 \\
\cline{2-5}
 & RoBERTa Base & 40.7 & 37.3 & 44.9 \\
\hline
\multirow{4}{*}{Objectives and Activities} & Longformer Base & 22.4 & 28.7 & 18.3 \\
\cline{2-5}
 & \textbf{LayoutLM-v3 Base} & \textbf{54.4} & \textbf{68.5} & \textbf{45.1} \\
\cline{2-5}
 & DeBERTa-v3 Base & 35.9 & 48.1 & 28.6 \\
\cline{2-5}
 & RoBERTa Base & 29.1 & 33.3 & 25.9 \\
\hline
\multirow{4}{*}{Principal Office City} & Longformer Base & 63.2 & 70.2 & 57.6 \\
\cline{2-5}
 & LayoutLM-v3 Base & 68.7 & 69.3 & 68.1 \\
\cline{2-5}
 & \textbf{DeBERTa-v3 Base} & \textbf{69.4} & 66.7 & \textbf{72.4} \\
\cline{2-5}
 & RoBERTa Base & 67.7 & \textbf{73.7} & 62.7 \\
\hline
\multirow{4}{*}{Principal Office Postal Code} & Longformer Base & 81.0 & 89.1 & 74.2 \\
\cline{2-5}
 & \textbf{LayoutLM-v3 Base} & \textbf{85.1} & 85.5 & 84.7 \\
\cline{2-5}
 & DeBERTa-v3 Base & 85.0 & 82.7 & \textbf{87.5} \\
\cline{2-5}
 & RoBERTa Base & 83.5 & \textbf{91.8} & 76.5 \\
\hline
\multirow{4}{*}{Principal Office Street Address} & Longformer Base & 64.5 & 69.6 & 60.0 \\
\cline{2-5}
 & LayoutLM-v3 Base & 57.6 & 58.9 & 56.4 \\
\cline{2-5}
 & \textbf{DeBERTa-v3 Base} & \textbf{71.4} & 67.0 & \textbf{76.5} \\
\cline{2-5}
 & RoBERTa Base & 65.1 & \textbf{75.0} & 57.5 \\
\hline
\multirow{4}{*}{Project Name} & Longformer Base & 5.1 & 3.0 & \textbf{16.7} \\
\cline{2-5}
 & \textbf{LayoutLM-v3 Base} & \textbf{16.5} & \textbf{24.2} & 12.5 \\
\cline{2-5}
 & DeBERTa-v3 Base & 7.0 & 6.1 & 8.3 \\
\cline{2-5}
 & RoBERTa Base & 13.6 & 12.1 & 15.4 \\
\hline
\multirow{4}{*}{Trustee Name} & Longformer Base & 49.1 & 39.6 & \textbf{64.4} \\
\cline{2-5}
 & \textbf{LayoutLM-v3 Base} & \textbf{50.5} & \textbf{41.7} & 64.0 \\
\cline{2-5}
 & DeBERTa-v3 Base & 45.1 & 36.4 & 59.0 \\
\cline{2-5}
 & RoBERTa Base & 49.3 & 40.3 & 63.4 \\
\hline
\multirow{4}{*}{Trustee Title} & Longformer Base & 69.1 & 65.4 & \textbf{73.2} \\
\cline{2-5}
 & \textbf{LayoutLM-v3 Base} & \textbf{72.0} & \textbf{71.7} & 72.3 \\
\cline{2-5}
 & DeBERTa-v3 Base & 66.7 & 62.1 & 71.9 \\
\cline{2-5}
 & RoBERTa Base & 69.2 & 69.4 & 68.9 \\
\hline
\multirow{4}{*}{Year Ended} & \textbf{Longformer Base} & \textbf{95.6} & \textbf{96.4} & 94.9 \\
\cline{2-5}
 & LayoutLM-v3 Base & 95.3 & 93.9 & \textbf{96.7} \\
\cline{2-5}
 & DeBERTa-v3 Base & 94.8 & 95.8 & 93.7 \\
\cline{2-5}
 & RoBERTa Base & 94.6 & 95.5 & 93.6 \\
\hline
\end{tabular}}\caption{Field level metrics for Charities Continued.}\end{table*}

\begin{table*}\centering\scriptsize{\begin{tabular}{|c|c|c|c|c|}
\hline
\textbf{Field Name} & \textbf{Base Model} & \textbf{Test F1} & \textbf{Test Precision} & \textbf{Test Recall} \\
\hline
\hline
\multirow{4}{*}{Advertiser} & Longformer Base & 49.1 & 45.9 & 52.9 \\
\cline{2-5}
 & \textbf{LayoutLM-v3 Base} & \textbf{59.3} & \textbf{60.9} & 57.8 \\
\cline{2-5}
 & DeBERTa-v3 Base & 57.6 & 56.8 & \textbf{58.4} \\
\cline{2-5}
 & RoBERTa Base & 49.6 & 48.2 & 51.2 \\
\hline
\multirow{4}{*}{Agency} & Longformer Base & 62.0 & 64.5 & \textbf{59.8} \\
\cline{2-5}
 & \textbf{LayoutLM-v3 Base} & \textbf{65.1} & \textbf{71.7} & 59.6 \\
\cline{2-5}
 & DeBERTa-v3 Base & 61.4 & 67.1 & 56.7 \\
\cline{2-5}
 & RoBERTa Base & 58.4 & 67.8 & 51.2 \\
\hline
\multirow{4}{*}{Agency Commission} & Longformer Base & 77.8 & 84.4 & 72.2 \\
\cline{2-5}
 & LayoutLM-v3 Base & 81.4 & \textbf{88.3} & 75.6 \\
\cline{2-5}
 & \textbf{DeBERTa-v3 Base} & \textbf{85.0} & \textbf{88.3} & \textbf{81.9} \\
\cline{2-5}
 & RoBERTa Base & 82.5 & 85.7 & 79.5 \\
\hline
\multirow{4}{*}{Gross Total} & \textbf{Longformer Base} & \textbf{78.0} & \textbf{86.8} & \textbf{70.8} \\
\cline{2-5}
 & LayoutLM-v3 Base & 73.4 & 79.9 & 67.9 \\
\cline{2-5}
 & DeBERTa-v3 Base & 71.9 & 80.5 & 65.0 \\
\cline{2-5}
 & RoBERTa Base & 72.1 & 77.4 & 67.6 \\
\hline
\multirow{4}{*}{Line Item - Days} & Longformer Base & 67.9 & 70.0 & 65.9 \\
\cline{2-5}
 & \textbf{LayoutLM-v3 Base} & \textbf{70.7} & 72.9 & \textbf{68.6} \\
\cline{2-5}
 & DeBERTa-v3 Base & 69.4 & \textbf{74.6} & 64.8 \\
\cline{2-5}
 & RoBERTa Base & 68.8 & 70.9 & 66.7 \\
\hline
\multirow{4}{*}{Line Item - Description} & Longformer Base & 49.2 & 76.2 & 36.3 \\
\cline{2-5}
 & LayoutLM-v3 Base & 49.3 & 74.3 & 36.9 \\
\cline{2-5}
 & \textbf{DeBERTa-v3 Base} & \textbf{50.9} & \textbf{79.5} & \textbf{37.4} \\
\cline{2-5}
 & RoBERTa Base & 49.1 & 77.1 & 36.1 \\
\hline
\multirow{4}{*}{Line Item - End Date} & \textbf{Longformer Base} & \textbf{93.3} & 95.4 & \textbf{91.2} \\
\cline{2-5}
 & LayoutLM-v3 Base & 92.5 & \textbf{98.1} & 87.6 \\
\cline{2-5}
 & DeBERTa-v3 Base & 90.5 & 96.8 & 84.9 \\
\cline{2-5}
 & RoBERTa Base & 91.8 & \textbf{98.1} & 86.2 \\
\hline
\multirow{4}{*}{Line Item - Rate} & Longformer Base & 50.0 & 62.3 & \textbf{41.8} \\
\cline{2-5}
 & LayoutLM-v3 Base & 47.0 & 59.3 & 38.9 \\
\cline{2-5}
 & \textbf{DeBERTa-v3 Base} & \textbf{50.1} & \textbf{65.2} & 40.7 \\
\cline{2-5}
 & RoBERTa Base & 48.4 & 62.3 & 39.5 \\
\hline
\multirow{4}{*}{Line Item - Start Date} & Longformer Base & 73.5 & 96.8 & 59.3 \\
\cline{2-5}
 & \textbf{LayoutLM-v3 Base} & \textbf{77.1} & \textbf{98.2} & \textbf{63.5} \\
\cline{2-5}
 & DeBERTa-v3 Base & 72.9 & 97.8 & 58.1 \\
\cline{2-5}
 & RoBERTa Base & 72.7 & 97.6 & 57.9 \\
\hline
\multirow{4}{*}{Net Amount Due} & Longformer Base & 72.4 & \textbf{84.7} & 63.3 \\
\cline{2-5}
 & LayoutLM-v3 Base & 74.0 & 78.2 & \textbf{70.3} \\
\cline{2-5}
 & \textbf{DeBERTa-v3 Base} & \textbf{75.5} & \textbf{84.7} & 68.2 \\
\cline{2-5}
 & RoBERTa Base & 70.6 & 82.3 & 61.8 \\
\hline
\multirow{4}{*}{Payment Terms} & Longformer Base & 67.2 & 82.0 & 56.9 \\
\cline{2-5}
 & LayoutLM-v3 Base & 61.8 & 67.0 & 57.3 \\
\cline{2-5}
 & \textbf{DeBERTa-v3 Base} & \textbf{75.7} & \textbf{84.0} & \textbf{68.9} \\
\cline{2-5}
 & RoBERTa Base & 67.2 & \textbf{84.0} & 56.0 \\
\hline
\end{tabular}}\caption{Field level metrics for FCC Invoices.}\end{table*}

\begin{table*}\centering\scriptsize{\begin{tabular}{|c|c|c|c|c|}
\hline
\textbf{Field Name} & \textbf{Base Model} & \textbf{Test F1} & \textbf{Test Precision} & \textbf{Test Recall} \\
\hline
\hline
\multirow{4}{*}{Effective Date} & Longformer Base & 75.5 & 78.0 & 73.2 \\
\cline{2-5}
 & LayoutLM-v3 Base & 74.7 & \textbf{81.3} & 69.2 \\
\cline{2-5}
 & \textbf{DeBERTa-v3 Base} & \textbf{81.6} & 80.2 & \textbf{83.0} \\
\cline{2-5}
 & RoBERTa Base & 76.2 & 79.1 & 73.5 \\
\hline
\multirow{4}{*}{Jurisdiction} & Longformer Base & 92.3 & 95.7 & 89.1 \\
\cline{2-5}
 & \textbf{LayoutLM-v3 Base} & \textbf{94.8} & 96.8 & \textbf{92.9} \\
\cline{2-5}
 & DeBERTa-v3 Base & 93.8 & 95.7 & 91.8 \\
\cline{2-5}
 & RoBERTa Base & 92.5 & \textbf{97.9} & 87.6 \\
\hline
\multirow{4}{*}{Party} & Longformer Base & 75.1 & 72.3 & 78.2 \\
\cline{2-5}
 & LayoutLM-v3 Base & 72.6 & 70.9 & 74.4 \\
\cline{2-5}
 & \textbf{DeBERTa-v3 Base} & \textbf{75.9} & 69.5 & \textbf{83.6} \\
\cline{2-5}
 & RoBERTa Base & 75.8 & \textbf{74.2} & 77.5 \\
\hline
\end{tabular}}\caption{Field level metrics for NDA.}\end{table*}

\begin{table*}\centering\scriptsize{\begin{tabular}{|c|c|c|c|c|}
\hline
\textbf{Field Name} & \textbf{Base Model} & \textbf{Test F1} & \textbf{Test Precision} & \textbf{Test Recall} \\
\hline
\hline
\multirow{4}{*}{(Header) Contract Area Description} & Longformer Base & 48.1 & 54.3 & 43.1 \\
\cline{2-5}
 & \textbf{LayoutLM-v3 Base} & \textbf{50.0} & \textbf{56.5} & 44.8 \\
\cline{2-5}
 & DeBERTa-v3 Base & 47.7 & \textbf{56.5} & 41.3 \\
\cline{2-5}
 & RoBERTa Base & 48.5 & 52.2 & \textbf{45.3} \\
\hline
\multirow{4}{*}{(Header) Environmental protections} & \textbf{Longformer Base} & \textbf{71.7} & 73.1 & \textbf{70.4} \\
\cline{2-5}
 & LayoutLM-v3 Base & 63.6 & 65.4 & 61.8 \\
\cline{2-5}
 & DeBERTa-v3 Base & 70.9 & \textbf{75.0} & 67.2 \\
\cline{2-5}
 & RoBERTa Base & 59.5 & 69.2 & 52.2 \\
\hline
\multirow{4}{*}{(Header) Governing law} & Longformer Base & 70.0 & \textbf{70.0} & 70.0 \\
\cline{2-5}
 & LayoutLM-v3 Base & 66.0 & \textbf{70.0} & 62.5 \\
\cline{2-5}
 & \textbf{DeBERTa-v3 Base} & \textbf{70.7} & \textbf{70.0} & \textbf{71.4} \\
\cline{2-5}
 & RoBERTa Base & 69.3 & \textbf{70.0} & 68.6 \\
\hline
\multirow{4}{*}{(Header) Hardship clause or force majeure} & Longformer Base & 81.8 & 83.7 & 80.0 \\
\cline{2-5}
 & LayoutLM-v3 Base & 81.4 & 81.4 & 81.4 \\
\cline{2-5}
 & \textbf{DeBERTa-v3 Base} & \textbf{84.1} & \textbf{86.0} & \textbf{82.2} \\
\cline{2-5}
 & \textbf{RoBERTa Base} & \textbf{84.1} & \textbf{86.0} & \textbf{82.2} \\
\hline
\multirow{4}{*}{(Header) Income tax: rate} & Longformer Base & 62.2 & 60.5 & \textbf{63.9} \\
\cline{2-5}
 & \textbf{LayoutLM-v3 Base} & \textbf{64.1} & \textbf{65.8} & 62.5 \\
\cline{2-5}
 & DeBERTa-v3 Base & 59.7 & 60.5 & 59.0 \\
\cline{2-5}
 & RoBERTa Base & 57.5 & \textbf{65.8} & 51.0 \\
\hline
\multirow{4}{*}{(Header) Reporting requirements} & \textbf{Longformer Base} & \textbf{65.7} & \textbf{67.0} & 64.5 \\
\cline{2-5}
 & LayoutLM-v3 Base & 62.2 & 65.1 & 59.5 \\
\cline{2-5}
 & DeBERTa-v3 Base & 65.4 & 64.2 & \textbf{66.7} \\
\cline{2-5}
 & RoBERTa Base & 58.4 & 60.4 & 56.6 \\
\hline
\multirow{4}{*}{(Header) Term} & Longformer Base & 61.7 & \textbf{68.3} & 56.2 \\
\cline{2-5}
 & LayoutLM-v3 Base & 54.5 & 65.0 & 47.0 \\
\cline{2-5}
 & \textbf{DeBERTa-v3 Base} & \textbf{64.8} & 58.3 & \textbf{72.9} \\
\cline{2-5}
 & RoBERTa Base & 58.6 & 65.0 & 53.4 \\
\hline
\multirow{4}{*}{(Header) Water use} & \textbf{Longformer Base} & \textbf{57.1} & 50.0 & 66.7 \\
\cline{2-5}
 & LayoutLM-v3 Base & 44.4 & 33.3 & 66.7 \\
\cline{2-5}
 & DeBERTa-v3 Base & 44.4 & \textbf{66.7} & 33.3 \\
\cline{2-5}
 & RoBERTa Base & 52.6 & 41.7 & \textbf{71.4} \\
\hline
\multirow{4}{*}{Contract Area Description} & Longformer Base & 14.7 & 16.8 & 13.1 \\
\cline{2-5}
 & \textbf{LayoutLM-v3 Base} & \textbf{19.0} & \textbf{19.8} & \textbf{18.2} \\
\cline{2-5}
 & DeBERTa-v3 Base & 9.9 & 10.9 & 9.0 \\
\cline{2-5}
 & RoBERTa Base & 7.6 & 10.9 & 5.8 \\
\hline
\multirow{4}{*}{Country} & \textbf{Longformer Base} & \textbf{61.6} & \textbf{67.0} & 57.0 \\
\cline{2-5}
 & LayoutLM-v3 Base & 54.9 & 58.2 & 52.0 \\
\cline{2-5}
 & DeBERTa-v3 Base & 61.1 & 63.7 & 58.6 \\
\cline{2-5}
 & RoBERTa Base & 59.3 & 59.3 & \textbf{59.3} \\
\hline
\multirow{4}{*}{Date Signed} & \textbf{Longformer Base} & \textbf{61.2} & \textbf{67.3} & \textbf{56.1} \\
\cline{2-5}
 & LayoutLM-v3 Base & 55.3 & 61.8 & 50.0 \\
\cline{2-5}
 & DeBERTa-v3 Base & 48.3 & 63.6 & 38.9 \\
\cline{2-5}
 & RoBERTa Base & 48.6 & 47.3 & 50.0 \\
\hline
\multirow{4}{*}{Environmental protections} & Longformer Base & 17.1 & \textbf{23.4} & 13.4 \\
\cline{2-5}
 & LayoutLM-v3 Base & 17.4 & \textbf{23.4} & 13.8 \\
\cline{2-5}
 & \textbf{DeBERTa-v3 Base} & \textbf{23.5} & 22.4 & \textbf{24.7} \\
\cline{2-5}
 & RoBERTa Base & 16.3 & 21.5 & 13.1 \\
\hline
\multirow{4}{*}{Governing law} & Longformer Base & 39.3 & \textbf{42.3} & 36.7 \\
\cline{2-5}
 & LayoutLM-v3 Base & 30.6 & 32.7 & 28.8 \\
\cline{2-5}
 & \textbf{DeBERTa-v3 Base} & \textbf{43.1} & \textbf{42.3} & \textbf{44.0} \\
\cline{2-5}
 & RoBERTa Base & 34.7 & 40.4 & 30.4 \\
\hline
\multirow{4}{*}{Hardship clause or force majeure} & Longformer Base & 29.9 & 38.5 & 24.4 \\
\cline{2-5}
 & LayoutLM-v3 Base & 27.9 & 36.5 & 22.6 \\
\cline{2-5}
 & \textbf{DeBERTa-v3 Base} & \textbf{34.1} & \textbf{42.3} & \textbf{28.6} \\
\cline{2-5}
 & RoBERTa Base & 16.2 & 23.1 & 12.5 \\
\hline
\multirow{4}{*}{Income tax: rate} & Longformer Base & 7.1 & 7.5 & 6.7 \\
\cline{2-5}
 & LayoutLM-v3 Base & 6.7 & 7.5 & 6.1 \\
\cline{2-5}
 & \textbf{DeBERTa-v3 Base} & \textbf{13.2} & \textbf{15.0} & \textbf{11.8} \\
\cline{2-5}
 & RoBERTa Base & 6.7 & 7.5 & 6.0 \\
\hline
\multirow{4}{*}{Participants} & Longformer Base & 53.1 & \textbf{57.8} & 49.1 \\
\cline{2-5}
 & LayoutLM-v3 Base & 47.7 & 54.2 & 42.5 \\
\cline{2-5}
 & \textbf{DeBERTa-v3 Base} & \textbf{54.3} & 54.7 & \textbf{53.9} \\
\cline{2-5}
 & RoBERTa Base & 48.0 & 53.3 & 43.6 \\
\hline
\multirow{4}{*}{Project} & Longformer Base & 35.3 & 36.7 & 34.0 \\
\cline{2-5}
 & \textbf{LayoutLM-v3 Base} & \textbf{38.4} & 38.8 & \textbf{38.0} \\
\cline{2-5}
 & DeBERTa-v3 Base & 30.8 & \textbf{44.9} & 23.4 \\
\cline{2-5}
 & RoBERTa Base & 28.9 & 24.5 & 35.3 \\
\hline
\multirow{4}{*}{Renewal or extension of term} & Longformer Base & 14.9 & \textbf{22.9} & 11.1 \\
\cline{2-5}
 & LayoutLM-v3 Base & 5.1 & 8.3 & 3.7 \\
\cline{2-5}
 & \textbf{DeBERTa-v3 Base} & \textbf{15.0} & 18.8 & \textbf{12.5} \\
\cline{2-5}
 & RoBERTa Base & 13.9 & 21.9 & 10.1 \\
\hline
\end{tabular}}\caption{Field level metrics for Resource Contracts.}\end{table*}

\begin{table*}\centering\scriptsize{\begin{tabular}{|c|c|c|c|c|}
\hline
\textbf{Field Name} & \textbf{Base Model} & \textbf{Test F1} & \textbf{Test Precision} & \textbf{Test Recall} \\
\hline
\hline
\multirow{4}{*}{Reporting requirements} & Longformer Base & 28.5 & \textbf{36.2} & 23.5 \\
\cline{2-5}
 & LayoutLM-v3 Base & 22.3 & 28.2 & 18.5 \\
\cline{2-5}
 & \textbf{DeBERTa-v3 Base} & \textbf{32.4} & 33.9 & \textbf{31.0} \\
\cline{2-5}
 & RoBERTa Base & 16.5 & 23.2 & 12.9 \\
\hline
\multirow{4}{*}{Signatories, company} & \textbf{Longformer Base} & \textbf{59.0} & \textbf{58.8} & \textbf{59.1} \\
\cline{2-5}
 & LayoutLM-v3 Base & 52.3 & 55.2 & 49.7 \\
\cline{2-5}
 & DeBERTa-v3 Base & 53.8 & 57.6 & 50.5 \\
\cline{2-5}
 & RoBERTa Base & 52.0 & 50.9 & 53.2 \\
\hline
\multirow{4}{*}{Term} & Longformer Base & 17.8 & 22.0 & 15.0 \\
\cline{2-5}
 & LayoutLM-v3 Base & 6.6 & 8.5 & 5.4 \\
\cline{2-5}
 & DeBERTa-v3 Base & 18.7 & \textbf{24.4} & 15.2 \\
\cline{2-5}
 & \textbf{RoBERTa Base} & \textbf{20.3} & \textbf{24.4} & \textbf{17.4} \\
\hline
\multirow{4}{*}{Type} & Longformer Base & 59.0 & 56.9 & 61.2 \\
\cline{2-5}
 & \textbf{LayoutLM-v3 Base} & \textbf{63.9} & 65.3 & \textbf{62.7} \\
\cline{2-5}
 & DeBERTa-v3 Base & 54.0 & \textbf{70.8} & 43.6 \\
\cline{2-5}
 & RoBERTa Base & 50.7 & 52.8 & 48.7 \\
\hline
\multirow{4}{*}{Water use} & Longformer Base & 30.8 & 32.3 & 29.4 \\
\cline{2-5}
 & LayoutLM-v3 Base & 26.9 & 29.0 & 25.0 \\
\cline{2-5}
 & \textbf{DeBERTa-v3 Base} & \textbf{49.2} & \textbf{48.4} & \textbf{50.0} \\
\cline{2-5}
 & RoBERTa Base & 32.8 & 32.3 & 33.3 \\
\hline
\end{tabular}}\caption{Field level metrics for Resource Contracts Continued.}\end{table*}

\begin{table*}\centering\scriptsize{\begin{tabular}{|c|c|c|c|c|}
\hline
\textbf{Field Name} & \textbf{Base Model} & \textbf{Test F1} & \textbf{Test Precision} & \textbf{Test Recall} \\
\hline
\hline
\multirow{4}{*}{(Header) Description of Securities} & Longformer Base & 79.5 & 87.0 & 73.2 \\
\cline{2-5}
 & LayoutLM-v3 Base & 80.0 & 87.0 & 74.1 \\
\cline{2-5}
 & \textbf{DeBERTa-v3 Base} & \textbf{87.9} & \textbf{89.9} & \textbf{86.1} \\
\cline{2-5}
 & RoBERTa Base & 82.2 & 87.0 & 77.9 \\
\hline
\multirow{4}{*}{(Header) Dividend Policy} & Longformer Base & 98.4 & \textbf{98.4} & 98.4 \\
\cline{2-5}
 & LayoutLM-v3 Base & 98.4 & \textbf{98.4} & 98.4 \\
\cline{2-5}
 & \textbf{DeBERTa-v3 Base} & \textbf{99.2} & \textbf{98.4} & \textbf{100.0} \\
\cline{2-5}
 & RoBERTa Base & 96.8 & 96.8 & 96.8 \\
\hline
\multirow{4}{*}{(Header) Prospectus Summary} & \textbf{Longformer Base} & \textbf{100.0} & \textbf{100.0} & \textbf{100.0} \\
\cline{2-5}
 & LayoutLM-v3 Base & 95.5 & 96.9 & 94.0 \\
\cline{2-5}
 & DeBERTa-v3 Base & 99.2 & \textbf{100.0} & 98.5 \\
\cline{2-5}
 & RoBERTa Base & 86.6 & 89.2 & 84.1 \\
\hline
\multirow{4}{*}{(Header) Risks To The Business} & Longformer Base & 75.0 & 94.0 & 62.4 \\
\cline{2-5}
 & LayoutLM-v3 Base & 67.0 & \textbf{95.5} & 51.6 \\
\cline{2-5}
 & DeBERTa-v3 Base & 90.0 & 94.0 & 86.3 \\
\cline{2-5}
 & \textbf{RoBERTa Base} & \textbf{92.5} & 92.5 & \textbf{92.5} \\
\hline
\multirow{4}{*}{Agent Address} & Longformer Base & 89.6 & 92.3 & 87.0 \\
\cline{2-5}
 & LayoutLM-v3 Base & 90.2 & 92.3 & 88.2 \\
\cline{2-5}
 & DeBERTa-v3 Base & 91.0 & \textbf{93.8} & 88.4 \\
\cline{2-5}
 & \textbf{RoBERTa Base} & \textbf{93.1} & \textbf{93.8} & \textbf{92.4} \\
\hline
\multirow{4}{*}{Agent Name} & Longformer Base & 77.9 & 77.3 & 78.5 \\
\cline{2-5}
 & \textbf{LayoutLM-v3 Base} & \textbf{84.6} & \textbf{83.3} & \textbf{85.9} \\
\cline{2-5}
 & DeBERTa-v3 Base & 78.2 & 78.8 & 77.6 \\
\cline{2-5}
 & RoBERTa Base & 82.2 & 80.3 & 84.1 \\
\hline
\multirow{4}{*}{Agent Telephone} & Longformer Base & 98.4 & 98.4 & 98.4 \\
\cline{2-5}
 & LayoutLM-v3 Base & 99.2 & \textbf{100.0} & 98.5 \\
\cline{2-5}
 & \textbf{DeBERTa-v3 Base} & \textbf{100.0} & \textbf{100.0} & \textbf{100.0} \\
\cline{2-5}
 & RoBERTa Base & 99.2 & 98.4 & \textbf{100.0} \\
\hline
\multirow{4}{*}{Amount Registered} & Longformer Base & 75.9 & 76.1 & 75.7 \\
\cline{2-5}
 & \textbf{LayoutLM-v3 Base} & \textbf{79.0} & \textbf{79.4} & \textbf{78.6} \\
\cline{2-5}
 & DeBERTa-v3 Base & 74.1 & 72.2 & 76.0 \\
\cline{2-5}
 & RoBERTa Base & 74.3 & 76.1 & 72.5 \\
\hline
\multirow{4}{*}{Attorney Names} & Longformer Base & 63.1 & 55.5 & 73.2 \\
\cline{2-5}
 & \textbf{LayoutLM-v3 Base} & \textbf{64.6} & \textbf{56.6} & \textbf{75.1} \\
\cline{2-5}
 & DeBERTa-v3 Base & 43.9 & 35.9 & 56.4 \\
\cline{2-5}
 & RoBERTa Base & 63.8 & 55.9 & 74.5 \\
\hline
\multirow{4}{*}{Company Address} & Longformer Base & 92.3 & 92.3 & \textbf{92.3} \\
\cline{2-5}
 & LayoutLM-v3 Base & 90.2 & 92.3 & 88.2 \\
\cline{2-5}
 & \textbf{DeBERTa-v3 Base} & \textbf{92.4} & \textbf{93.8} & 91.0 \\
\cline{2-5}
 & RoBERTa Base & 92.3 & 92.3 & \textbf{92.3} \\
\hline
\multirow{4}{*}{Company Name} & Longformer Base & 94.7 & \textbf{94.0} & 95.5 \\
\cline{2-5}
 & \textbf{LayoutLM-v3 Base} & \textbf{95.5} & \textbf{94.0} & \textbf{96.9} \\
\cline{2-5}
 & DeBERTa-v3 Base & 93.2 & 92.5 & 93.9 \\
\cline{2-5}
 & \textbf{RoBERTa Base} & \textbf{95.5} & \textbf{94.0} & \textbf{96.9} \\
\hline
\multirow{4}{*}{Company Officer} & Longformer Base & 87.2 & 85.1 & 89.4 \\
\cline{2-5}
 & \textbf{LayoutLM-v3 Base} & \textbf{88.9} & \textbf{86.3} & \textbf{91.7} \\
\cline{2-5}
 & DeBERTa-v3 Base & 86.0 & 81.1 & \textbf{91.7} \\
\cline{2-5}
 & RoBERTa Base & 86.2 & 83.2 & 89.6 \\
\hline
\multirow{4}{*}{Company Officer Title} & Longformer Base & 93.5 & 90.8 & 96.2 \\
\cline{2-5}
 & \textbf{LayoutLM-v3 Base} & \textbf{93.9} & \textbf{91.5} & \textbf{96.5} \\
\cline{2-5}
 & DeBERTa-v3 Base & 90.0 & 87.1 & 93.1 \\
\cline{2-5}
 & RoBERTa Base & 92.0 & 90.2 & 93.9 \\
\hline
\multirow{4}{*}{Date of Prospectus} & \textbf{Longformer Base} & \textbf{96.9} & \textbf{95.5} & \textbf{98.4} \\
\cline{2-5}
 & \textbf{LayoutLM-v3 Base} & \textbf{96.9} & \textbf{95.5} & \textbf{98.4} \\
\cline{2-5}
 & \textbf{DeBERTa-v3 Base} & \textbf{96.9} & \textbf{95.5} & \textbf{98.4} \\
\cline{2-5}
 & \textbf{RoBERTa Base} & \textbf{96.9} & \textbf{95.5} & \textbf{98.4} \\
\hline
\multirow{4}{*}{Description of Securities (1st Para)} & Longformer Base & 53.3 & 74.3 & 41.6 \\
\cline{2-5}
 & \textbf{LayoutLM-v3 Base} & \textbf{73.0} & \textbf{77.1} & \textbf{69.2} \\
\cline{2-5}
 & DeBERTa-v3 Base & 54.0 & 72.9 & 42.9 \\
\cline{2-5}
 & RoBERTa Base & 56.4 & 75.7 & 44.9 \\
\hline
\multirow{4}{*}{Dividend Policy (1st Para)} & \textbf{Longformer Base} & \textbf{88.2} & 88.9 & \textbf{87.5} \\
\cline{2-5}
 & LayoutLM-v3 Base & 81.9 & \textbf{93.7} & 72.8 \\
\cline{2-5}
 & DeBERTa-v3 Base & 76.6 & 85.7 & 69.2 \\
\cline{2-5}
 & RoBERTa Base & 64.7 & 87.3 & 51.4 \\
\hline
\multirow{4}{*}{EIN} & Longformer Base & 97.6 & 96.9 & \textbf{98.4} \\
\cline{2-5}
 & LayoutLM-v3 Base & 97.7 & \textbf{98.4} & 96.9 \\
\cline{2-5}
 & \textbf{DeBERTa-v3 Base} & \textbf{98.4} & \textbf{98.4} & \textbf{98.4} \\
\cline{2-5}
 & RoBERTa Base & 96.9 & 96.9 & 96.9 \\
\hline
\multirow{4}{*}{Joint Book Runners} & Longformer Base & 52.5 & 43.1 & 67.1 \\
\cline{2-5}
 & \textbf{LayoutLM-v3 Base} & \textbf{53.6} & \textbf{44.0} & \textbf{68.6} \\
\cline{2-5}
 & DeBERTa-v3 Base & 48.6 & 38.5 & 65.6 \\
\cline{2-5}
 & RoBERTa Base & 51.4 & 42.2 & 65.7 \\
\hline
\end{tabular}}\caption{Field level metrics for S1.}\end{table*}

\begin{table*}\centering\scriptsize{\begin{tabular}{|c|c|c|c|c|}
\hline
\textbf{Field Name} & \textbf{Base Model} & \textbf{Test F1} & \textbf{Test Precision} & \textbf{Test Recall} \\
\hline
\hline
\multirow{4}{*}{Law Firm Address} & \textbf{Longformer Base} & \textbf{69.2} & \textbf{60.7} & \textbf{80.3} \\
\cline{2-5}
 & LayoutLM-v3 Base & 67.1 & 59.5 & 76.9 \\
\cline{2-5}
 & DeBERTa-v3 Base & 66.7 & 58.9 & 76.7 \\
\cline{2-5}
 & RoBERTa Base & 68.0 & 59.5 & 79.4 \\
\hline
\multirow{4}{*}{Law Firm Name} & Longformer Base & 79.0 & \textbf{75.8} & 82.5 \\
\cline{2-5}
 & \textbf{LayoutLM-v3 Base} & \textbf{80.0} & \textbf{75.8} & \textbf{84.7} \\
\cline{2-5}
 & DeBERTa-v3 Base & 76.1 & 71.8 & 80.9 \\
\cline{2-5}
 & RoBERTa Base & 78.3 & 74.2 & 82.9 \\
\hline
\multirow{4}{*}{Max Price} & Longformer Base & 60.6 & 57.0 & 64.8 \\
\cline{2-5}
 & \textbf{LayoutLM-v3 Base} & \textbf{64.6} & \textbf{61.0} & \textbf{68.5} \\
\cline{2-5}
 & DeBERTa-v3 Base & 51.9 & 56.0 & 48.3 \\
\cline{2-5}
 & RoBERTa Base & 57.0 & 59.0 & 55.1 \\
\hline
\multirow{4}{*}{Prospectus Summary (1st Para)} & Longformer Base & 76.1 & 77.2 & 75.1 \\
\cline{2-5}
 & LayoutLM-v3 Base & 76.2 & 77.4 & 75.0 \\
\cline{2-5}
 & \textbf{DeBERTa-v3 Base} & \textbf{88.1} & \textbf{91.7} & \textbf{84.9} \\
\cline{2-5}
 & RoBERTa Base & 75.3 & 77.7 & 73.0 \\
\hline
\multirow{4}{*}{Risk Clauses} & \textbf{Longformer Base} & \textbf{96.7} & 97.0 & 96.4 \\
\cline{2-5}
 & LayoutLM-v3 Base & 96.5 & 96.2 & \textbf{96.8} \\
\cline{2-5}
 & DeBERTa-v3 Base & 93.7 & 95.6 & 91.9 \\
\cline{2-5}
 & RoBERTa Base & 94.9 & \textbf{97.1} & 92.8 \\
\hline
\multirow{4}{*}{Title of Security Registered} & Longformer Base & 86.3 & 87.2 & 85.3 \\
\cline{2-5}
 & \textbf{LayoutLM-v3 Base} & \textbf{89.6} & \textbf{90.6} & \textbf{88.6} \\
\cline{2-5}
 & DeBERTa-v3 Base & 86.6 & 86.1 & 87.1 \\
\cline{2-5}
 & RoBERTa Base & 83.8 & 85.0 & 82.7 \\
\hline
\end{tabular}}\caption{Field level metrics for S1 Continued.}\end{table*}

\end{document}

\appendix
\maketitle

\section{Dataset Information}
\label{sec:dataset-information-appendix}

In this section we detail the intended use and possible misuse of our data as well as licenses for the original documents and the labels we are releasing. 

The bucket URIs below are references to Wasabi buckets and can be downloaded with the following command
\begin{verbatim}
  aws s3 sync <source> <destination> --endpoint-url=https://s3.us-east-2.wasabisys.com --no-sign-request
\end{verbatim}

Note that Croissant files only contain text and labels, for rich OCR formats, images or original documents please see the full dataset at the Download Locations below. For more information on data sources, data volumes and label fields, please refer to the paper.

The datasets are also available on Zenodo to ensure long term availability \url{https://zenodo.org/records/13327077}.

\textbf{Intended Uses - }
The intended use of these datasets is to benchmark models for Key Information Extraction on documents and tasks that contain realistic challenges as seen in an industrial setting. The output of such work is intended to be the techniques \textbf{not} the models trained on these datasets. Some examples of this work may include:
\begin{itemize}
    \item Development and testing of improved layout-aware models capable of improving on beyond the performance of text-only models for datasets with complex tables such as our FCC Invoices dataset
    \item Development and testing of techniques for handling long documents, including longer-context models applicable to KIE tasks, or improved chunking techniques for existing models.
    \item Development and testing of techniques for information extraction from noisy text data, such as robust tokenization or data augmentation techniques.
\end{itemize}

\textbf{Potential for Misuse - }
In the authors opinion, potential for misuse falls into two main categories:
\begin{itemize}
    \item \textbf{Application to Production Systems:} Deploying a model trained on these datasets in a real-world application. These datasets are not intended to be production-ready, are not licensed for commercial use, and may not cover all edge cases and variability found in real-world documents.
    \item \textbf{Extraction of Personal Information: } There is potential for these datasets to be used to train models that extract personal information from documents, such as names, addresses, and other sensitive data. Users might attempt to leverage these models for tasks involving personal data extraction, increasing the risk of privacy violations and misuse of sensitive information. However, the risk is considered acceptable because existing Named Entity Recognition (NER) datasets are likely to be equally or more capable for such purposes.
\end{itemize}

\subsection{Charities}

\textbf{Bucket URI}: \nolinkurl{s3://project-fruitfly/charities} \\
\textbf{Croissant File}: \url{https://s3.us-east-2.wasabisys.com/project-fruitfly/charities/croissant.json} \\
\textbf{Label License}: CC-BY-NC 4.0 \\
\textbf{Document Source}: UK Charities Commission \url{https://register-of-charities.charitycommission.gov.uk/} \\
\textbf{Document License}: Open Government License - V2 \\

\subsection{FCC Invoices}

\textbf{Bucket URI}: \nolinkurl{s3://project-fruitfly/fcc_invoices} \\
\textbf{Croissant File}: \url{https://s3.us-east-2.wasabisys.com/project-fruitfly/fcc_invoices/croissant.json} \\
\textbf{Label License}: CC-BY-NC 4.0 \\
\textbf{Document Source}: FCC \url{https://publicfiles.fcc.gov/} \\
\textbf{Document License}: Fair use under 17 USC 107. At the time of writing, the FCC site states the portal's objective is to "make information to which the public already has a right more readily available, so that the public will be encouraged to play a more active part in dialogue with broadcast licensees."

\subsection{Resource Contracts}

\textbf{Bucket URI}: \nolinkurl{s3://project-fruitfly/resource_contracts} \\
\textbf{Croissant File}: \url{https://s3.us-east-2.wasabisys.com/project-fruitfly/resource_contracts/croissant.json} \\
\textbf{Label License}: CC-BY-NC 4.0 \\
\textbf{Document Source}: Resource Contracts \url{https://www.resourcecontracts.org/} \\
\textbf{Document License}: CC-BY-SA 4.0 \\

\subsection{S1}

\textbf{Bucket URI}: \nolinkurl{s3://project-fruitfly/s1_pages} \\
\textbf{Croissant File}: \url{https://s3.us-east-2.wasabisys.com/project-fruitfly/s1_pages/croissant.json} \\
\textbf{Label License}: CC-BY-NC 4.0 \\
\textbf{Document Source}: EDGAR (SEC) \url{https://www.sec.gov/edgar}\\
\textbf{Document License}: Subject to the dissemination section of the Edgar Website \url{https://www.sec.gov/privacy}. At time of collection this states. "Information presented on \nolinkurl{www.sec.gov} is considered public information and may be copied or further distributed by users of the web site without the SEC's permission."\\
\textbf{Document Level Files}: A document level version of the dataset, with the same splits is available at \nolinkurl{s3://project-fruitfly/s1} \\

\subsection{NDA}

\textbf{Bucket URI}: \nolinkurl{s3://project-fruitfly/nda} \\
\textbf{Croissant File}: \url{https://s3.us-east-2.wasabisys.com/project-fruitfly/nda/croissant.json} \\
\textbf{Label License}: CC-BY-NC 4.0 \\
\textbf{Document Source}: EDGAR (SEC) \url{https://www.sec.gov/edgar}\\
\textbf{PDF Source}: Kleister-NDA \url{https://github.com/applicaai/kleister-nda/}
\textbf{Document License}: Subject to the dissemination section of the Edgar Website \url{https://www.sec.gov/privacy}. At time of collection this states. "Information presented on \nolinkurl{www.sec.gov} is considered public information and may be copied or further distributed by users of the web site without the SEC's permission."\\

\section{Maintenance Plan}
The datasets will be stored long-term in a public bucket on Wasabi (an S3-like blob store) where they currently reside. In case any issues with Wasabi arise, we will switch to an alternative blob-store such as Amazon's S3 and update our Github page to reflect the new location of the files.

We currently have no plans to release updates to this dataset but may do so in the event of data quality issues or other issues brought to our attention by the community.  Minimizing changes to this dataset over time helps to ensure a stable baseline on which to evaluate future modeling work.

\section{Data Formats}
For each dataset, the data is pre-split into 3 csv files: train.csv, test.csv and val.csv. Each of the CSVs has the following columns.

\textbf{text:} This is the raw document text output from the OCR engine.\\
\textbf{labels:} Each document's labels are a list of JSON encoded objects each representing a single span. Each span object has a "start" and "end" field, indexing into the document text. A text field containing the text value of the label. and a "label" field which contains the label name. For example:
\begin{verbatim}
[
 {
  start: 10,
  end: 16,
  label: "cost",
  text: $10.99
 }, ...
]
\end{verbatim}
\textbf{ocr: } A path, relative to the root of the dataset directory. The path points to a gzipped JSON file containing the full document OCR output. An example of the format of the OCR is: \\
\begin{verbatim}
[
  {
    "pages": [
      {
        "text": ...,
        "doc_offset": {"start": 0, "end": 1234},
        "page_num": 0,
      }
    ],
    "blocks": [
      {
        "text": ...,
        "doc_offset": {"start": 0, "end": 1234},
        "position": {
          "top": 34,
          "bottom": 64,
          "left": 123,
          "right":246
        },
      },
      ...
    ],
    "tokens": [
      {
        "text": ...,
        "doc_offset": {"start": 0, "end": 1234},
        "position": {...},
      },
      ...
    ]
    "chars": [
      {
        "doc_index": 0,
        "block_index": 0,
        "page_index": 0,
        "confidence": 100,
        "page_num": 0,
        "position": {...},
        "text": "E"
      },
      ...
    ]
  },
  ...
]
\end{verbatim}
\textbf{document\_path: } A path, relative to the root of the dataset directory. The path points to the original pdf.\\
\textbf{image\_paths: } A JSON encoded object containing a list of paths. The paths point to a PNG representation of each of the pages.\\

\begin{figure}[H]
    \centering
\begin{python}
import gzip
import json
import os
import typing as t
import pandas as pd
from PIL import Image

def read_data(
  dataset_dir: str,
  dataset: t.Literal[
    "fcc_invoices",
    "nda",
    "charities",
    "s1_pages",
    "resource_contracts",
  ],
  split: t.Literal["train", "test", "val"],
):
  # Read the CSV
  df = pd.read_csv(os.path.join(dataset_dir, dataset, f"{split}.csv"))

  # Iterate over the rows
  for row in df.to_dict("records"):
    # Get the OCR Document text
    text = row["text"]

    # Get and deserialize the Labels
    # Formatted as above
    labels = json.loads(row["labels"])

    # Decompress and deserialize the OCR
    # Formatted as above
    with gzip.open(os.path.join(dataset_dir, row["ocr"]), "rt") as fp:
      ocr = json.load(fp)

    # Read in Images. One per page
    page_images = []
    for page_image_path in json.loads(row["image_files"]):
      page_images.append(
        Image.open(os.path.join(dataset_dir, page_image_path))
      )

    # The document path
    document_pdf_path = os.path.join(dataset_dir, row["document_path"])

    yield text, labels, ocr, page_images, document_pdf_path
\end{python}
    \caption{Example Python code for loading the data, including text, labels, images, OCR and document paths}
    \label{fig:code}
\end{figure}

\section{Author statement}
The authors of this paper bear all responsibility in case of violation of rights, etc. associated with the RealKIE dataset.